\documentclass{article} % For LaTeX2e
\usepackage{iclr2020_conference,times}
\PassOptionsToPackage{square, numbers, compress,sort}{natbib}

\usepackage{graphicx}
\usepackage{subfigure}
\usepackage{wrapfig}
\usepackage{booktabs}
\usepackage{amsmath,amsfonts,bm}
\usepackage[utf8]{inputenc} % allow utf-8 input
\usepackage[T1]{fontenc}    % use 8-bit T1 fonts
\usepackage{hyperref}       % hyperlinks
\usepackage{url}            % simple URL typesetting
\usepackage{amsfonts}       % blackboard math symbols
\usepackage{nicefrac}       % compact symbols for 1/2, etc.
\usepackage{graphicx}
\usepackage{subfigure}
\usepackage{dsfont}
\usepackage{xspace}
\usepackage{amssymb}
\usepackage{mathtools}
\usepackage{hyperref}
\usepackage{amsfonts}
\usepackage{color}

\usepackage{float}
\usepackage{scalerel,stackengine}
\usepackage{soul}
\usepackage[algo2e,ruled,vlined]{algorithm2e}
\include{pythonlisting}
\usepackage{appendix}
\usepackage{amsthm}

\usepackage{wasysym}
\usepackage{pifont}
\usepackage{bbm}
\usepackage{lipsum}
\def\figref#1{figure~\ref{#1}}
\def\eqref#1{equation~\ref{#1}}
\def\1{\bm{1}}

\def\mW{{\bm{W}}}

\def\mSigma{{\bm{\Sigma}}}

\DeclareMathAlphabet{\mathsfit}{\encodingdefault}{\sfdefault}{m}{sl}
\SetMathAlphabet{\mathsfit}{bold}{\encodingdefault}{\sfdefault}{bx}{n}

\newcommand{\E}{\mathbb{E}}

\usepackage{amsmath}
\usepackage{bm}
\usepackage{hyperref}
\usepackage{url}
\usepackage{graphicx,multicol}
\DeclareFontFamily{OMS}{oasy}{\skewchar\font48 }
\DeclareFontShape{OMS}{oasy}{m}{n}{%
         <-5.5> oasy5     <5.5-6.5> oasy6
      <6.5-7.5> oasy7     <7.5-8.5> oasy8
      <8.5-9.5> oasy9     <9.5->  oasy10
      }{}
\DeclareFontShape{OMS}{oasy}{b}{n}{%
       <-6> oabsy5
      <6-8> oabsy7
      <8->  oabsy10
      }{}
\DeclareSymbolFont{oasy}{OMS}{oasy}{m}{n}
\SetSymbolFont{oasy}{bold}{OMS}{oasy}{b}{n}

\DeclareMathSymbol{\smallleftarrow}     {\mathrel}{oasy}{"20}
\DeclareMathSymbol{\smallrightarrow}    {\mathrel}{oasy}{"21}
\DeclareMathSymbol{\smallleftrightarrow}{\mathrel}{oasy}{"24}

\title{Probabilistic Numeric \\ Convolutional Neural Networks}

\author{Marc Finzi\thanks{Work done during internship at Qualcomm AI Research} \\
Qualcomm AI Research \\ New York University\\
\texttt{maf820@nyu.edu} \\
\And
Roberto Bondesan \& Max Welling \\
Qualcomm AI Research\thanks{Qualcomm AI Research is an initiative of Qualcomm Technologies, Inc.} \\
\texttt{\{rbondesa, mwelling\}@qti.qualcomm.com}\\
}

\iclrfinalcopy % Uncomment for camera-ready version, but NOT for submission.
\begin{document}

\maketitle

\begin{abstract}
Continuous input signals like images and time series that are irregularly sampled or have missing values are challenging for existing deep learning methods. Coherently defined feature representations must depend on the values in unobserved regions of the input. Drawing from the work in probabilistic numerics, we propose Probabilistic Numeric Convolutional Neural Networks which represent features as Gaussian processes (GPs), providing a probabilistic description of discretization error. We then define a convolutional layer as the evolution of a PDE defined on this GP, followed by a nonlinearity. This approach also naturally admits steerable equivariant convolutions under e.g.~the rotation group. In experiments we show that our approach yields a $3\times$ reduction of error from the previous state of the art on the SuperPixel-MNIST dataset and competitive performance on the medical time series dataset PhysioNet2012. 
\end{abstract}

\section{Introduction}

Standard convolutional neural networks are defined on a regular input grid. For continuous signals like time series and images, these elements correspond to regular samples of an underlying function $f$ defined on a continuous domain.
%: \mathbb{R}^d \to \mathbb{R}^c$ where $c$ is the number of channels and $d$ is the dimension of the input space. 
In this case, the standard convolutional layer of a neural network is a numerical approximation of a continuous convolution operator $\mathcal{A}$.

%A typical convolutional neural network is a sequence of these layers, interspersed with pointwise nonlinearities. Increasingly more expressive representations are built up recursively. 
Coherently defined networks on continuous functions should \emph{only} depend on the input function $f$, and not on spurious shortcut features \citep{geirhos2020shortcut} such as the sampling locations or sampling density, which enable overfitting and reduce robustness to changes in the sampling procedure.
% \begin{equation}
%     \mathrm{NN} = \mathcal{A}^{(L)}\circ\mathrm{ReLU} \circ\mathcal{A}^{(L-1)}\circ ...\circ \mathcal{A}^{(2)}\circ\mathrm{ReLU} \circ\mathcal{A}^{(1)}
% \end{equation}
Each application of $\mathcal{A}$ in a standard neural network incurs some discretization error which is determined by the sampling resolution. In some sense, this error is unavoidable because the features $f^{(\ell)}$ at the layers $\ell$ depend on the values of the input function $f$ at regions that \emph{have not been observed}.
For input signals which are sampled at a low resolution, or even sampled irregularly such as with the sporadic measurements of patient vitals data in ICUs or dispersed sensors for measuring ocean currents, this discretization error cannot be neglected. Simply filling in the missing data with zeros or imputing the values is not sufficient since many different imputations are possible, each of which can affect the outcomes of the network.

Probabilistic numerics is an emergent field that studies discretization errors in numerical algorithms using probability theory \cite{cockayne2019bayesian}. Here we build upon these ideas to quantify the dependence of the network on the regions in the input which are unknown, and integrate this uncertainty into the computation of the network. To do so, we replace the discretely evaluated feature maps $\{f^{(\ell)}(x_i)\}_{i=1}^N$ with Gaussian processes: distributions over the continuous function $f^{(\ell)}$ that track the most likely values as well as the uncertainty. 
On this Gaussian process feature representation, we need not resort to discretizing the convolution operator $\mathcal{A}$ as in a standard convnet, but instead we can apply the continuous convolution operator directly. If a given feature  is a Gaussian process, then applying linear operators yields a new Gaussian process with transformed mean and covariance functions. 
%If we know the mean and covariance functions of the input, we can compute the mean and covariance of the output which has been transformed by a linear operator. 
The dependence of $\mathcal{A}f$ on regions of $f$ which are not known translates into the uncertainty represented in the transformed covariance function, the analogue of the discretization error in a CNN, which is now tracked explicitly.
We call the resulting model Probalistic Numeric Convolutional Neural Network (PNCNN).

% In the following sections, we describe how these parametrized translation equivariant linear operators can be constructed as the solution to the anisotropic diffusion equation, how they can be applied in closed form to RBF-kernel GPs, and how the $\mathrm{ReLU}$ nonlinearity transforms both the mean and the covariance function.
% Finally, we will present numerical experiments on irregularly sampled data validating the usefulness of our approach.

\section{Related Work}
Over the years there have been many successful convolutional approaches for ungridded data such as GCN \citep{kipf2016semi}, PointNet \citep{qi2017pointnet}, Transformer \citep{vaswani2017attention}, Deep Sets \citep{zaheer2017deep}, SplineCNN \citep{fey2018splinecnn}, PCNN \citep{atzmon2018point}, PointConv \citep{wu2019pointconv}, KPConv \citep{thomas2019kpconv} and many others \citep{de2020gauge,finzi2020generalizing,schutt2017schnet,wang2018deep}. However, the target domains of sets, graphs, and point clouds are intrinsically discrete and for continuous data each of these methods fail to take full advantage of the assumption that the underlying signal is continuous. Furthermore, none of these approaches reason about the underlying signal probabilistically. 
%\MW{Maybe we should also cite mesh-CNNs as they do try to approximate a continuous signal on a manifold with a graph and then pass messages on that graph. (BTW: it would be very cool to generalize mesh-CNNs to PN-MESH-CNNs! We would need equivariant GP kernels to handle gauge transformations.)}

In a separate line of work there are several approaches tackling irregularly spaced time series with RNNs \citep{che2018recurrent}, Neural ODEs \citep{rubanova2019latent}, imputation to a regular grid \citep{li2016scalable,futoma2017learning,shukla2019interpolation,fortuin2020gp}, set functions \citep{horn2019set} and attention \citep{shukla2020multi}. Additionally there are several works exploring reconstruction of images from incomplete observations for downstream classification \citep{huijben2019deep,li2020learning}.

Most similar to our method are the end-to-end Gaussian process adapter \citep{li2016scalable} and the multi-task Gaussian process RNN classifier \citep{futoma2017learning}. In these two works, a Gaussian process is fit to an irregularly spaced time series and sampled imputations from this process are fed into a separate RNN classifier. Unlike our approach where the classifier operates directly on a continuous and probabilistic signal, in these works 
%the models for imputation and classification are separate entities with 
the classifier operates on a deterministic signal on a regular grid and cannot reason probabilistically about discretization errors.
%As a result, these approaches cannot integrate uncertainty in the intermediate layers or reason probabilistically about discretization errors.

Finally, while superficially similar to Deep GPs \citep{damianou2013deep} or Deep Differential Gaussian Process Flows \citep{hegde2018deep}, our PNCNNs tackle fundamentally different kinds of problems like image classification\footnote{While GPs \textit{can} be applied directly to image classification, they are not well suited to this task even with convolutional structure baked in, as shown in \citet{kumar2018deep}.}, and our GPs represent epistemic uncertainty over the values of the feature maps rather than the parameters of the network.

% Deep learning with differential Gaussian process flows
% \citep{hegde2018deep}
% Group Invariance, Stability to Deformations,
% and Complexity of Deep Convolutional Representations
% Probabilistic propagation of uncertainties through ReLUs and network:
% Lightweight Probabilistic Networks
% \citep{gast2018lightweight}
% \citep{kong2020sde}
% Derivatives:
% Deep neural networks motivated by partial differential equations,

%\section{Gaussian Processes and Linear PDEs}
\section{Background}\label{sec:background}
%\subsection{Probabilistic interpolation with Gaussian Processes}
\textbf{Probabilistic Numerics:} \hspace{1em}
We draw inspiration for our approach from the community of \textit{probabilistic numerics} where the error in numerical algorithms are modeled probabilistically, and typically with a Gaussian process. In this framework, only a finite number of input function calls can be made, and therefore the numerical algorithm can be viewed as an autonomous agent which has epistemic uncertainty over the values of the input. A well known example is Bayesian Monte Carlo where a Gaussian process is used to model the error in the numerical estimation of an integral and optimally select a rule for its computation \citep{minka2000deriving,rasmussen2003bayesian}.
 Probabilistic numerics has been applied widely to numerical problems such as the inversion of a matrix \citep{hennig2015probabilistic}, the solution of an ODE \citep{schober2019probabilistic}, a meshless solution to boundary value PDEs \citep{cockayne2016probabilistic}, and other numerical problems \citep{cockayne2019bayesian}. To our knowledge, we are the first to construct a probabilistic numeric method for convolutional neural networks.

% \subsection{Gaussian Processes}
% \label{sec:GP}
\textbf{Gaussian Processes:} \hspace{1em}
% We would like to define a neural network that operates on the underlying continuous function $f(x)$ where $x$ belongs to a continuous domain like $\mathbb{R}$ for the continuous time for time series or $\mathbb{R}^2$ for the continuous space of an image. In practice however, we have access only to a collection of the values of that function sampled on a finite number of points $x_i$, the setting of a probabilistic numerical algorithm. In general one cannot reconstruct the values of $f$ elsewhere, and without further structure there may be many functions consistent with the observations.
We are interested in operating on the continuous function $f(x)$ underlying the input, but in practice we have access only to a collection of the values of that function sampled on a finite number of points $\{x_i\}_{i=1}^N$. Classical interpolation theory reconstructs $f$ deterministically by assuming a certain structure of the signal in the frequency domain. Gaussian processes give a way of modeling our beliefs about values that have not been observed \citep{rasmussen2006gaussian}, as reviewed in appendix \ref{sec:GP_review}. These beliefs are encoded into a prior covariance $k$ of the GP $f\sim\mathcal{GP}(0,k)$ and updated upon seeing data with Bayesian inference.
Explicitly, given a set of sampling locations $\bm{x} = \{x_i\}_{i=1}^N$ and noisy observations $\bm{y} = \{y_i\}_{i=1}^N$ sampled $y_i \sim {\cal N}(f(x_i), \sigma_i^2)$, using Bayes rule one can compute the posterior distribution $f|\bm{y},\bm{x} \sim \mathcal{GP}(\mu_p, k_p)$, which captures our epistemic uncertainty about the values between observations. The posterior mean and covariance are given by
\begin{equation}
\label{eq:posterior}
\mu_p(x)=\bm{k}(x)^\top [K + S]^{-1} \bm{y}
\,,\quad
k_p(x,x')
=
k(x, x')
-
\bm{k}(x)^\top [K + S]^{-1}
\bm{k}(x')\,,
\end{equation}
where $K_{ij}=k(x_i,x_j), k(x)_i = k(x, x_i)$
and $S = \text{diag}(\sigma_i^2)$.
%The predictive distribution is then $p(y_*|x_*,\mY,\mX) = {\cal N}(\mu_p(x_*), k_p(x_*,x_*) + \sigma_*^2)$.
Below we shall choose the RBF kernel\footnote{For convenience, we include the additional scale factor ${(2\pi l^2)}^{d/2}$ relative to the usual definition.} as prior covariance, due to its convenient analytical properties:
% $k_{\mathrm{RBF}}(x,x') = 
%     a
%     \mathcal{N}(x;x',l^2I)
%     =
%     \frac{a}{{(2\pi l^2)}^{d/2}}
%     \exp(-\tfrac{1}{2 l^2}(x-x')^2)\,,
% $
%\begin{align}
%    \label{eq:RBF}
$    k_{\mathrm{RBF}}(x,x') = 
    a
    \mathcal{N}(x;x',l^2I)
    =
    a \left(2\pi l^2\right)^{-\frac{d}{2}}
    \exp(-\tfrac{1}{2 l^2}||x-x'||^2)$.
    %\,,
%\end{align}
%due to its convenient analytical properties w.r.t.~integration against another Gaussian kernel.
In typical applications of GPs to machine learning tasks such as regression, the function $f$ that we want to predict is already the regression model.
%each $x_i$ is a datapoint and the task is to predict the values of the function $f$ at some test points $x^*$. 
In contrast, here we use GPs as a way of representing our beliefs and epistemic uncertainty about the values of both the input function and the intermediate feature maps of a neural network.
%\RB{I find this confusing, remove?}\MW{I agree}Each example time series or image input is represented by a distinct Gaussian process and transformed by convolutional layers to other Gaussian process feature maps. 
% In typically applications of GPs to machine learning tasks, the model is applied to a regression or classification problem where the input $x$ to the function is already the signal one wants to regress or classify. We stress that here we use GPs to interpolate the underlying signal instead and we will use a neural network to map the GP interpolation to a class label or regression output as we shall discuss in the next section.
\section{Probabilistic Numeric Convolutional Neural Networks}
\begin{figure}
    \centering
    \includegraphics[width=\textwidth]{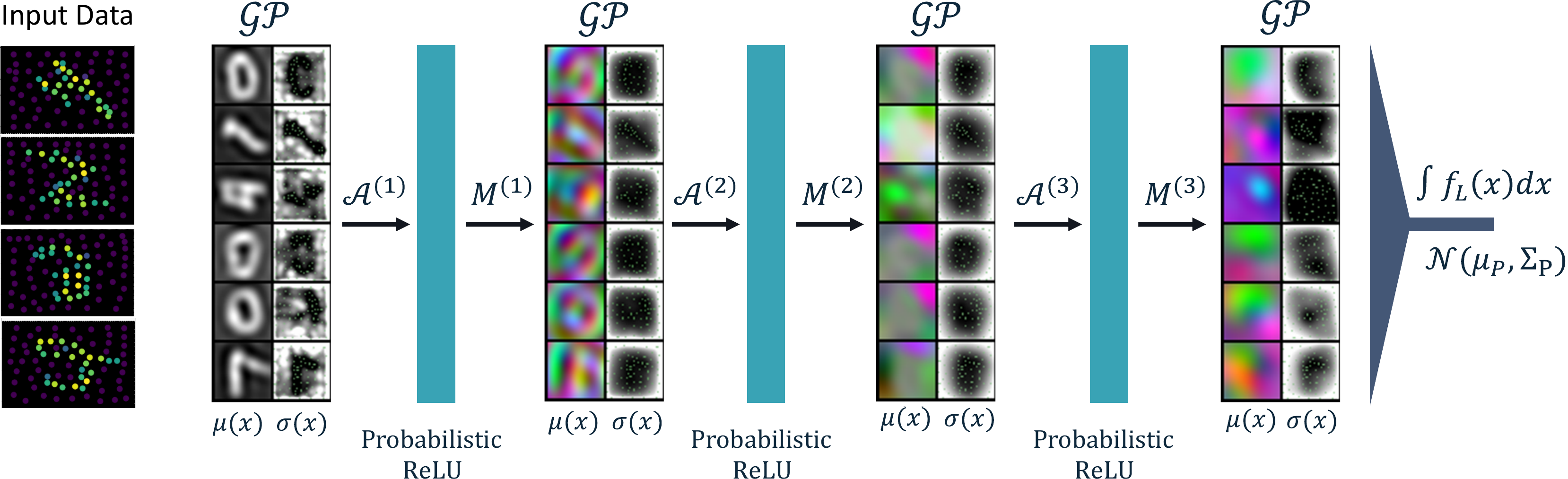}
    \vspace{-4mm}
    \caption{The PNCNN operating on SuperPixel-MNIST images shown on the left. The mean and elementwise uncertainty of the Gaussian process feature maps are shown as they are transformed through the network by the convolution layers. Observation points shown as green dots in $\sigma(x)$.
    }
    \label{fig:network}
    %\vspace{-5mm}
\end{figure}
\subsection{Overview}
Given an input signal $f : {\cal X} \to \mathbb{R}^c$, we define a network with layers that act directly on this continuous input signal. We define our neural network recursively from the input $f^{(0)}=f$, as a series of $L$ continuous convolutions $\mathcal{A}^{(\ell)}$ with pointwise ReLU nonlinearities and weight matrices which mix only channels (known as $1\times1$ convolutions) $M \in \mathbb{R}^{c\times c}$:
% \begin{align}
%     \Phi
%     = 
%     {\cal P}
%     \circ 
%     \sigma 
%     \circ 
%     {\cal A}^{(L)}
%     \cdots 
%     \circ 
%     \sigma
%     \circ 
%     {\cal A}^{(2)}
%     \circ 
%     \sigma
%     \circ 
%     {\cal A}^{(1)}    
%     \,.
% \end{align}
\begin{equation}\label{eq:layer_recurrence}
    f^{(\ell+1)} = M^{(\ell)}\mathrm{ReLU}[\mathcal{A}^{(\ell)}f^{(\ell)}],
\end{equation}
and a final global average pooling layer $\mathcal{P}$ which acts channel-wise as natural generalization of the discrete case: $\mathcal{P}(f^{(L)})_\alpha = \int f^{(L)}_\alpha(x){\rm d}x$ for each $\alpha=1,2,\dots,c$. %\footnote{Here Euclidean volume element ${\rm d}x$ is a stand-in for a measure ${\rm d}\mu(x)$ for more general spaces ${\cal X}$ such as a manifold or a discrete lattice. If $\mathcal{X}$ is discrete such as $\mathcal{X}=\mathbb{Z}^2$, one gets standard average pooling $\mu(x) = \sum_{z\in \mathbb{Z}^2} \delta(x-z)$.}
Denoting the space of functions on ${\cal X}$ with $c$ channels by ${\cal H}_c$, the convolution operators ${\cal A}^{(\ell)}$ are linear operators from ${\cal H}_{c_\ell}$ to ${\cal H}_{c_{\ell+1}}$. Like in ordinary convolutional neural networks, the layers build up increasingly more expressive spatial features and depend on the parameters in $\mathcal{A}^{(\ell)}$ and $M^{(\ell)}$. Unlike ordinary convolutional networks, these layers are well defined operations on the underlying continuous signal.

While it is clear that such a network can be defined abstractly, the exact values of the function $f^{(L)}$ cannot be computed as the operators depend on unknown values of the input. However, by adapting a probabilistic description we can formulate our ignorance of $f^{(0)}$ with a Gaussian process and see how the uncertainties propagate through the layers of the network, yielding a probabilistic output. %we can form the GP posterior given the collection of values which are observed: $\{f(x_i)\}_{i=1}^N$. 
% \MW{I propose to remove the next sentences:}This GP feature map tracks both the most likely value of $f(x)$ given by $\mu_p(x)$ along with our epistemic uncertainty in the value of $f(x)$ given by $k_p(x,x)$, which increases in the regions that have not been observed. 
% %Using a GP to represent our uncertainty in the values of $f^{(\ell)}$ in each of the layers, we can then compute how the uncertainty is transformed by computations through the layers of the network. 
% To overcome the intractability of the model, we approximate the non-Gaussian intermediate features with a GP, which leads to an approximate yet efficient procedure. \RB{Marc to rewrite}\MW{Until here}
Before delving into details, 
we outline the key components of \eqref{eq:layer_recurrence} that make this possible.

\textbf{Continuous Convolutional Layers:} Crucially, we consider continuous convolution operators $\mathcal{A}$ that can be applied to input Gaussian process $f \sim \mathcal{GP}(\mu_p,k_p)$ in closed form. The output is another Gaussian process with a transformed mean and covariance $\mathcal{A}f \sim \mathcal{GP}(\mathcal{A}\mu_p,\mathcal{A}k_p\mathcal{A}')$ where $\mathcal{A}'$ acts to the left on the primed argument of $k_p(x,x')$.\footnote{More generally neural networks have affine layers including both convolutions and biases. An affine transformation $\mathcal{A}f+b$ of a Gaussian process is also a Gaussian process $f\sim \mathcal{GP}(\mathcal{A}\mu_p+b,\mathcal{A}k_p\mathcal{A}')$, and we include biases in our network but omit them from the derivations for simplicity.} In section \ref{subsec:continuous_convolution} we show how to parametrize these continuous convolutions in terms of the flow of a PDE and show how they can be applied to the RBF kernel exactly in closed form.

\textbf{Probabilistic ReLUs:} Applying the $\mathrm{ReLU}$ nonlinearity to the GP yields a new non Gaussian stochastic process $h^{(\ell)}=\mathrm{ReLU}[\mathcal{A}^{(\ell)}f^{(\ell)}]$, and we show in section \ref{subsec:nonlinearity} that the mean and covariance of this process has a closed form solution which can be computed.

\textbf{Channel Mixing and Central Limit Theorem:} The activations $h^{(\ell)}$ are not Gaussian; however, for a large number of weakly dependent channels we argue that $f^{(\ell+1)} = M^{(\ell)}h^{(\ell)}$ is approximately distributed as a Gaussian Process in section \ref{subsec:nonlinearity}. 

\textbf{Measurement and Projection to RBF Gaussian Process:} While $f^{(\ell+1)}$ is approximately a Gaussian process, the mean and covariance functions have a complicated form. Instead of using these functions directly, we take measurements of the mean and variance of this process and feed them in as noisy observations to a fresh RBF kernel GP, allowing us to repeat the process and build up multiple layers without increasing complexity.

The Gaussian process feature maps in the final layer $f^{(L)}$ are aggregated spatially by the integral pooling $\mathcal{P}$ which can also be applied in closed form (see appendix \ref{sec:integral_pooling}), to yield a Gaussian output. Assembling these components, we implement the end to end trainable Probabilistic Numeric Convolutional Neural Network which integrates a probabilistic description of missing data and discretization error inherent to continuous signals. The layers of the network are shown in \figref{fig:network}.

\subsection{Continuous convolutional layers}\label{subsec:continuous_convolution}
On a discrete domain such as the lattice $\mathcal{X}=\mathbb{Z}^d$, all translation equivariant linear operators $\mathcal{A}$ are convolutions, a fact which we review in appendix \ref{sec:disc_cont_conv}. In general, these convolutions can be written in terms of a linear combination of powers of the generators of the translation group: the shift operators %\MW{there should be d shift operators?}
$\tau_i,i=1,\dots,d$ shift all elements by one unit along the $i$-th axis of the grid. For a one dimensional grid, one can always write $\mathcal{A}=\sum_k W_k \tau^k$ where the weight matrices $W_k\in\mathbb{R}^{c\times c}$ act only on the channels and the shift operator $\tau$ acts on functions on the lattice. In $d$ dimensions, $\mathcal{A}= \sum_{k_1,\dots,k_d}W_{k_1,\dots,k_d} \tau_1^{k_1}\cdots \tau_d^{k_d}$ for some set of integer coefficients $k_1,\dots,k_d$. For example when $d=2$, we can take $k_1,k_2\in \{-1,0,1\}$ to fill out a $3\times 3$ neighborhood.

On the continuous domain $\mathcal{X}=\mathbb{R}^d$ we similarly parametrize convolutions with $\mathcal{A}=\sum_kW_k{\rm e}^{\mathcal{D}_k}$, where $\mathcal{D}_k$ is given by powers of the
partial derivatives $\partial_i, i=1,\dots,d$ which generate infinitesimal translations along the $i$-th axes. Setting $d=1$ for simplicity, we can indeed verify by Taylor expansion that the operator exponential $\tau^a = {\rm e}^{a\partial}$ applied to a function $g(x)$ is a translation: ${\rm e}^{a\partial}g(x)
=
g(x)
+ 
ag'(x)
+
\tfrac{1}{2}
a^2g''(x)
+\dots
=
g(x+a)$. Exponentials of operators can be defined similarly in terms of formal Taylor expansions or more conveniently in terms of the solution to a PDE:
%The application of the exponential to an input function $f(x)$ is the solution to the PDE
\begin{align}
    \label{eq:pde}
    \partial_t g(t, x) = ({\cal D} g)(t, x)\,,\quad 
    g(0, x) = g(x) \,,
\end{align}
at time $t=1$: ${\rm e}^{\mathcal{D}}g(x) = g(t=1,x)$.

Following the discussion in the discrete case, translation invariance of $\mathcal{D}_k$ imposes that it is expressed in terms of powers of the generators. 
%For a one dimensional signal, one can write ${\cal D} = \sum_{i\ge 0} a_i \partial_x^i$ for arbitrary constant $a_i$'s. Moving to multiple dimensions and 
Collecting the derivatives into the gradient $\nabla$, we can write the general form of
${\cal D}_k$ as $\alpha_k+
    \beta_k^\top \nabla + \tfrac{1}{2}
    \nabla^\top \Sigma_k\nabla+...$ for any constants $\alpha_k$, vectors $\beta_k$, matrices $\Sigma_k$ etc. For simplicity, we truncate the series at second order to get 
\begin{equation}
    \label{eq:D_heat}
    {\cal D}_k=
    \beta_k^\top \nabla + \tfrac{1}{2}
    \nabla^\top \Sigma_k\nabla\,,
\end{equation}
where we omit the constants $\alpha_k$ that can be absorbed into the definition of $W_k$. For this choice of $\mathcal{D}$, the PDE in \eqref{eq:pde} is nothing but the diffusion equation with drift $\beta_k$ and diffusion $\Sigma_k$. When discussing rotational equivariance in section \ref{sec:equivariance}, we also consider a more general form of $\mathcal{D}$.

The diffusion layer can also be viewed in another way as the infinitesimal generator of an Ito diffusion (a stochastic process). Given an Ito process with constant drift and diffusion ${\rm d}X_t = \beta {\rm d}t + \Sigma^{1/2}{\rm d}B_t$ where $B_t$ is a $d$ dimensional Brownian motion, the time evolution operator can be written via the Feynman-Kac formula as
${\rm e}^{t\mathcal{D}}f(x) = \E[f(X_t)]$ where $X_0 = x$. In other words, the operator layer $\mathcal{A}={\rm e}^{t\mathcal{D}}$ is the expectation under a parametrized Neural Stochastic Differential equation \citep{li2020scalable,tzen2019neural} that is homogeneous and therefore shift invariant. The flow of this SDE depends on the drift and diffusion parameters $\beta$ and $\Sigma$. 

% By taking linear combinations of different values $\beta_k$ and $\Sigma_k$ we can form a linear operator that couples different channels $\alpha,\beta=1,2,...,c$ like in a discrete convolutional layer: $\mathcal{A}[f](x)^\alpha = \sum_\beta \mathcal{A}^{\alpha\beta}f^\beta:=\sum_{\beta,k} W^{\alpha\beta}_ke^{\mathcal{D}_k}f^{\beta}$ or more succinctly $\mathcal{A}f = \sum_{k} W_ke^{\mathcal{D}_k}f$ in matrix notation.
To recap, we define our convolution operator through the general form $\mathcal{A} = \sum_kW_k{\rm e}^{\mathcal{D}_k}$ where the weight matrices $W_k\in \mathbb{R}^{c\times c}$ mix only channels and ${\rm e}^{\mathcal{D}_k}$ is the forward evolution by one unit of time of the diffusion equation with drift $\beta_k$ and diffusion $\Sigma_k$ containing learnable parameters $\{(W_k,\beta_k,\Sigma_k)\}_{k=1}^K$. The translation equivariance of $\mathcal{A}$ follows directly from the fact that the generators commute $\forall k,i: [\mathcal{D}_k,\nabla_i]=0$ and therefore $[\mathcal{A},\tau_{i}]=0$.
In appendix \ref{sec:disc_cont_conv} we show that our definition of ${\cal A}$ reduces to the usual one in the discrete case and is thus a principled generalization to the continuous domain.% and we compare the action of a differential and exponential operator on an input GP in appendix \ref{sec:linear_op_comparison}.

% Final piece the linear operator, use RBF.
\subsection{Exact Application on RBF GPs}
\label{sec:rbf_conv}

%\subsubsection{Operator Exponential and Greens Functions}
% The derivative is an entirely local operation. In the absence of additional blurring layers (see Section \ref{subsec:blurring}), it has a receptive field of $0$. For that reason, we may instead want to consider the solution of the PDE generated by a given differential operator. More precisely, given a differential operator $\mathcal{D}$ (local) we may consider its exponential \footnote{One can define the operator exponential formally by the Taylor series, but in fact it can be applied to non smooth functions through the definition as the (weak) solution to the PDE} $e^{t\mathcal{D}}$  which has a spatial extent. $e^{t\mathcal{D}}f(x)$ is the solution of the PDE $\frac{\partial f(x,t)}{\partial t}=\mathcal{D}f(x,t)$ at time $t$ with initial condition $f(x,0) = f(x)$.
Although the application of the linear operator $\mathcal{A} = \sum_kW_k{\rm e}^{\mathcal{D}_k}$ involves the time evolution of a PDE, owing to properties of the RBF kernel we fortuitously can apply the operator to an input GP in closed form!
Gaussian processes are closed under linear transformations: given $f\sim {\cal GP}(\mu_p, k_p)$, we need only compute the action of $\mathcal{A}$ on the mean and covariance: ${\cal A}f \sim {\cal GP}({\cal A}\mu_p, {\cal A}k_p{\cal A}')$, where ${\cal A}'$ is the adjoint w.r.t.~the $L_2({\cal X})$ inner product. 
The application of time evolution ${\rm e}^{\mathcal{D}_k}$ is a convolution with a Green's function $G_k$, so $\mathcal{A}f = \sum_kW_k{\rm e}^{\mathcal{D}_k}f = \sum_kW_kG_k*f$. As we derive in appendix \ref{subsec:greens}, the Green's function for $\mathcal{D}_k = \beta^\top_k \nabla + (1/2)\nabla^\top \Sigma_k \nabla$, is nothing but the multivariate Gaussian density $G_k(x) = {\cal N}(x;-\beta_k,\Sigma_k)$:
%{\rm e}^{-\frac{1}{2}(x+\beta_k)^\top\Sigma_k^{-1}(x+\beta_k)}\mathrm{det}(2\pi \Sigma_k)^{-1/2}$. %This implies that the mean of the transformed GP is:
%\MW{Here the function if a deterministic right? I find it confusing as we we use f for a random function and we should refer back to eqn 13 how one applies a linear operator on a random function.}
\begin{equation}
    \mathcal{A}f = \sum_{k}W_k{\rm e}^{\mathcal{D}_k}f
    =
\sum_{k}W_kG_k*f = \sum_{k}W_k\mathcal{N}(-\beta_k,\Sigma_k)*f\,.
\end{equation} 
%This operator can be equivalently expressed using the Feynman-Kac formula as $e^{t\mathcal{A}}f(x) = e^{t\alpha}\E[f(X_t)]$ with the Ito diffusion $dX_t = \beta dt + \Sigma^{1/2}dW_t$ starting at $X_0=x$. 
% To make further progress we note that we can apply these operators in closed form if the GP is interpolating data as  \eqref{eq:posterior} with $k=k_{\text{RBF}}$, 
In order to apply ${\rm e}^{t\mathcal{D}}$ to the posterior GP, we need only to be able to apply the operator to the posterior mean and covariance. This posterior mean and covariance in \eqref{eq:posterior} are expressed in terms of $k_{\text{RBF}}=a\mathcal{N}(x;x',\ell^2I)$ and the computation boils down to a convolution of two Gaussians: 
% We specialize now to the posterior process of \eqref{eq:posterior}
%with $k_{\text{RBF}}=a\mathcal{N}(x;x',\ell^2I)$.
%, in order 
%to apply $e^{t\mathcal{D}}$ onto a GP, we need to the posterior mean $\mu_p$ and covariance $k_p$, we need only be able to compute its application on $k_{\text{RBF}}$ which appears in both, and therefore 
%Then the computation boils down to a convolution of two Gaussians: 
%\MW{Do we need the first of these two equations? The action of A on the mean is similar to the eqn above and the action on K is the last of these two equations. Where is the first of these two used?}\marc{The one sided RBF application comes up in both the posterior mean and posterior covariance, I guess that also means we should try and make it clearer that the RBF comes up in the mean.}
\begin{align}
    \label{eq:A_rbf}
    {\rm e}^{t\mathcal{D}}k_{\mathrm{RBF}}(x,x') &= \mathcal{N}(x; -t\beta,t\Sigma)*a\mathcal{N}(x;x',\ell^2 I) 
    = a\mathcal{N}(x; x'-t\beta,\ell^2 I+ t\Sigma)
    \\
    \label{eq:A_rbf_A}
    {\rm e}^{t\mathcal{D}_1}k_{\mathrm{RBF}}(x,x'){\rm e}^{t\mathcal{D}_2'} 
    &= a\mathcal{N}(x; x'-t(\beta_1-\beta_2),\ell^2 I+t\Sigma_1+t\Sigma_2)\,.
\end{align}
The application of the channel mixing matrices $W_k$ and summation is also straightforward through matrix multiplication for the mean and covariance.
% We form our operator layer (mixing the channels) as
% $\mathcal{A} = \sum_i W_i e^{\mathcal{D}_i}$ in terms of a set of operators $\mathcal{A}_i = \beta_i^\top \nabla +(1/2)\nabla^\top \Sigma_i \nabla$ depending on the learned coefficients $\beta_i$ and $\Sigma_i$, coupled to matrices $W_i$ which mix the channels like in the derivative case and with normal convolutions:
% \begin{equation}
%     \mathcal{B}[f](x)^\alpha = \sum_\beta \mathcal{A}^{\alpha \beta}f(x)^\beta = \sum_{\beta,i} W_i^{\alpha \beta} e^{\mathcal{D}_i} f(x)^\beta.
% \end{equation}
%Here we have absorbed the scaling coefficients $\alpha_i$ into the weight matrices $K_i$. This layer contains the learnable parameters both for the channel mixing (like in the derivative case) and also additional learnable parameters $\{\beta_i,\Sigma_i\}_{i=1}^c$ for the generator (which become separate parameters because of the exponential's nonlinear dependence on $\beta$ and $\Sigma$). 
To summarize, because of the closed form action on the RBF kernel, the layer can be implemented efficiently and \textbf{exactly} with \textbf{no discretization} or approximations.
%\MW{I would write this section as: we now want to compute eqn 13 for a GP with an RBF kernel and a operator A as follows .... (derive stuff) The final closed form (!) expression is thus given by: (stuff). In general we need to build the story so that people keep seeing what we are doing in the big picture, by referring backwards and explaining forwards what we will do next.}
% \RB{The following part is unclear. There is some discussion of relation ordinary conv in section 3. Also, I remember that in fact Sigma=0 does not give reduced performance?}\MW{I kind of really like this limiting case, because it connects back to the classical results.}

We note that with the Green's function above, the action of ${\cal A}$ encompasses the ordinary convolution operator on the $2$d lattice as a special case. Given drift $\beta_k \in\{-1,0,1\}^{\times 2}, k=1,\dots,9$ filling out the $9$ elements of a $3\times 3$ grid and as the diffusion $\Sigma_k \to 0$, the Green's function is a Dirac delta, so that: $\mathcal{A}f(x) = \sum_{k}W_{k}\delta(x-\beta_k)*f(x) =\sum_{i,j=-1,0,1}W_{ij}f(x_1-i,x_2-j) =  W*_{\mathbb{Z}^2}f(x)$. %While having $\Sigma=0$ is a reasonable on a discrete grid where there is no meaning to intermediate function values, for continuous functions we need to be insensitive to changes in the input function of measure $0$ and therefore need to integrate the values over finite regions of space. 
%\marc{discuss relationship to anisotropic diffusion and scale spaces}

% Finally, we show in appendix \ref{sec:pooling} that the global average pooling action can be also computed exactly for an interpolating GP with RBF kernel by performing a Gaussian integral.

\subsection{General equivariance}
\label{sec:equivariance}
% \MW{We should write one paragraph and move the rest to the appendix. The paper os strong enough as is.}
The convolutional layers discussed so far are translation equivariant. %We now comment on the extension to equivariance under 
We discuss how to extend the continuous linear operator layers to more general symmetries such as rotations. Feature fields in this more general case are described by tensor fields, where the symmetry group acts not only on the input space ${\cal X}$ but also on the vector space attached to each point $x\in {\cal X}$. A linear layer ${\cal A}$ is equivariant if its action commutes with that of the symmetry.
In appendix \ref{sec:equivariance_app} we derive constraints for general linear operators and symmetries, which generalize those appearing in the steerable-CNN literature \citep{weiler2019general,cohen2019general}. Then we
show how equivariance under continuous roto-translations in 2d constrains the form of a convolutional layer by solving the equivariance constraint. Non-trivial solutions require that the operator $\mathcal D$ in the PDE of \eqref{eq:pde} has a non-trivial matrix structure. 

\subsection{Probabilistic Nonlinearities and Rectified Gaussian Processes}\label{subsec:nonlinearity}
%We leave the numerical study of the performance of such more general equivariant models for future work.

%\MW{Could move equations to the appendix and briefly mention that you can compute these things analytically}
%Just like how Gaussians transformed through nonlinear functions are no longer Gaussian, the activations $h^{(\ell)}=\mathrm{ReLU}[\mathcal{A}^{(\ell)}f^{(\ell)}]$ are not distributed as a Gaussian process. Instead they are distributed

\citet{gast2018lightweight} derive the mean and variance for a univariate rectified Gaussian distribution for use in a neural network. We generalize these results to the full covariance function (and higher moments) of a rectified Gaussian process in appendix \ref{sec:relu_covar} and present the results here. For the input GP $\mathcal{A}^{(\ell)}f^{(\ell)}(x)\sim {\cal GP}(\mu(x), k(x,x'))$, we denote $\sigma(x) = \sqrt{k(x,x)}$, $\mSigma$ the matrix with components $\Sigma_{ij}=k(x_i,x_j)$ for $i,j=1,2$ and $\bm{\mu}=[\mu(x_1),\mu(x_2)]$. We use notation $\Phi(z)$ for the univariate standard normal CDF, and $\bm{\Phi}(\bm{z};\mSigma)$ for (two dimensional) multivariate CDF of ${\cal N}(0,\mSigma)$ at $\bm{z}$. $\bm{\Sigma}_1$ and $\bm{\Sigma}_2$ are 
the column vectors of $\bm{\Sigma}$. The first and second moments of $h=\mathrm{ReLU}[\mathcal{A}f]$ are:
\begin{align}\label{eq:rectified_moments_mu}
    &\mathbb{E}[h(x)] 
    = \mu(x)\Phi(\mu(x)/\sigma(x)) + \sigma(x)\Phi'(\mu(x)/\sigma(x))
    \,,\\
    &\mathbb{E}
    [h(x_1)h(x_2))]
    = 
    (k(x_1,x_2)+  \mu(x_1) \mu(x_2)) \bm{\Phi}(\bm{\mu}; \mSigma)
    \label{eq:rectified_moments_Sigma}
    \\
    &\qquad
    + 
    (\mu(x_1)\bm{\Sigma}_{2}^\top  + \mu(x_2)\bm{\Sigma}_{1}^\top)\nabla\bm{\Phi}(\bm{\mu}; \mSigma)
    + \bm{\Sigma}_1^\top\nabla \nabla^\top\bm{\Phi}(\bm{\mu}; \mSigma)\bm{\Sigma}_2
    .\nonumber
\end{align}
The first and higher order derivatives of the Normal CDF are just the PDF and products of the PDF with Hermite polynomials. Note that the mean and covariance interact through the nonlinearity.

\subsection{Channel Mixing and Central Limit Theorem}

% While we can compute the mean and covariance of the rectified Gaussian process, the expressions (\ref{eq:rectified_moments_mu}, \ref{eq:rectified_moments_Sigma}) assume that the input is also a Gaussian process. If we want to coherently use this formulation over multiple layers, we need to transform approximately back to a Gaussian process, i.e. make the higher order cumulants as small as possible. 
% We address this challenge 
%We address the challenge of non-Gaussianity due to non-linearities.
 %In order for the modeling assumption that $f^{(\ell+1)}$ be distributed as a GP be not that egregious, we will add an additional linear layer that is pointwise over space but mixes the channels with a matrix $W$.
After the non-linearity the process is no longer Gaussian. To overcome this issue %we introduce an additional channel mixing %layer.
%Given the input function $f^{(\ell)}$ which is a Gaussian process (with multiple channels), 
we introduce a channel mixing 
matrix $M^{(\ell)}\in \mathbb{R}^{c_{\ell+1} \times c_\ell}$ and define the feature map in the following layer by $f^{(\ell+1)} = M^{(\ell)} h^{(\ell)}$, where $h^{(\ell)}=\mathrm{ReLU}[\mathcal{A}^{(\ell)}f^{(\ell)}]$. 
So long as the channels of $h^{(\ell)}$ are only weakly dependent, we can apply the central limit theorem (CLT) to each function 
$f^{(\ell+1)}_\alpha = \sum_{\beta=1}^{c_\ell}M_{\alpha,\beta}^{(\ell)} h^{(\ell)}_\beta$ so that in the limit of large $c_\ell$, the statistics of the
$f_\alpha^{(\ell+1)}$'s converge to a GP with first and second moments given by:
\begin{equation}\label{eq:mixing}
    \mathbb{E}[f^{(\ell+1)}(x)]
    = 
    M
    \mathbb{E}[h^{(\ell)}(x)] 
    ,
    \quad 
    \mathbb{E}[f^{(\ell+1)}(x)
    f^{(\ell+1)}(x')^\top] 
    = 
    M
    \mathbb{E}[h^{(\ell)}(x)
    h^{(\ell)}(x')^\top] 
    M^{\top}
    \,.
\end{equation}
We expand the argument in more detail in appendix \ref{sec:clt} and we quantify the extent of convergence and normality in the next section.
%figure \ref{fig:calibration} 
%\RB{Which? Isn't this already covered in next section?}.%since a stochastic process is defined by its finite distributions, and for each of those the CLT holds true.
% The extent to which the independence assumption is violated is studied in the appendix. \RB{reference QQ plot}

\subsection{Measurement and Projection to RBF Gaussian Process}\label{sec:projection}

As a last step we simplify the mean and covariance functions of the approximate GP $f^{(\ell+1)}$. While we can readily compute the values of these functions, unlike in the RBF kernel case, we cannot apply the convolution operator ${\rm e}^{t\mathcal{D}}$ in closed form. In order to circumvent this challenge, we model the (approximately) Gaussian process $f^{(\ell+1)}$ with an RBF Gaussian process as follows: we evaluate the mean $y_i=\mathbb{E}[f^{(\ell+1)}(x_i)]$ and variance $\sigma^2_i = \mathbb{V}\mathrm{ar}[f^{(\ell+1)}(x_i)]$ of the approximate Gaussian process $f^{(\ell+1)}$ at a collection of points $\{x_i\}_{i=1}^N$ using equations \ref{eq:rectified_moments_mu}, \ref{eq:rectified_moments_Sigma} and \ref{eq:mixing}. These values $y_i$ are treated as measurements of the underlying signal with a heteroscedastic noise $\sigma_i^2$ that varies from point to point. We can then compute the RBF-based posterior GP of this signal $\hat{f}^{(\ell+1)}|\{(x_i,y_i,\sigma_i)\}_{i=1}^N \sim \mathcal{GP}(\mu_p,k_p)$ with posterior mean and covariance given by \eqref{eq:posterior} for the heteroschedastic noise model. The uncertainty in the input $f^{(\ell)}$ is propagated through to the RBF posterior $\hat{f}^{(\ell+1)}|\{(x_i,y_i,\sigma_i)\}_{i=1}^N$ via the measurement noise $\sigma_i$. Crucially, this Gaussian process mean and covariance functions are written in terms of the RBF kernel and we can therefore continue applying convolutions in closed form in future layers.

\begin{figure}
\includegraphics[width=1.0\linewidth]{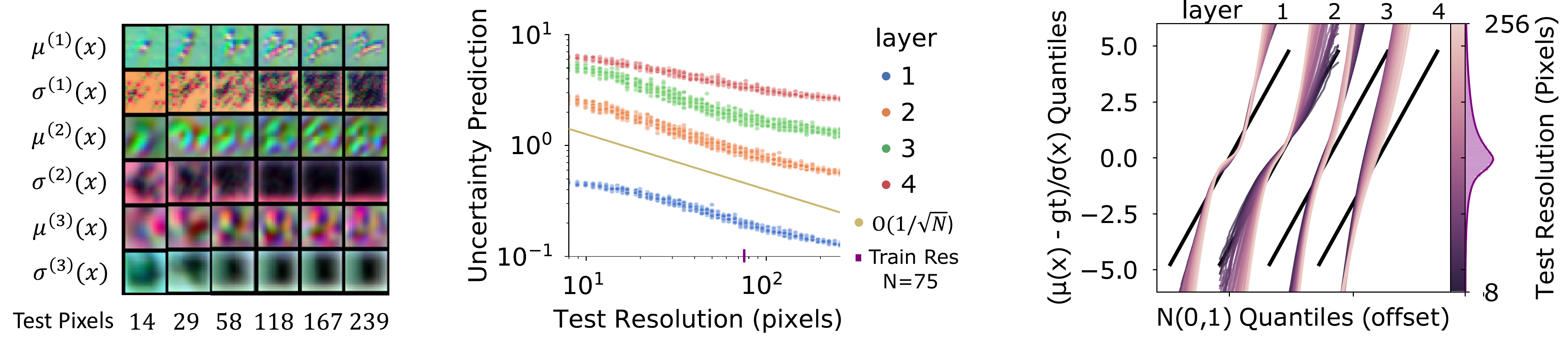}\\
% \begin{multicols}{3}
% \raggedcolumns
% \centering
% \includegraphics[width=.8\linewidth]{figs/feat_convergence.pdf}\\
% %\hspace{-.2\linewidth}
% \includegraphics[width=1.2\linewidth]{figs/uncscaling1.pdf}\\

% \includegraphics[width=1.0\linewidth]{figs/QQ_labeled.pdf}\\

% \end{multicols}

\caption{\textbf{Left:} Qualitative convergence of the mean and uncertainty of the first $3$ channels of the feature maps is shown in RGB color as the input test resolution is increased. \textbf{Middle:} Median predicted uncertainties over spatial locations as a function of the test resolution. \textbf{Right:} Using the predictions of the highest resolution model as ground truth, the distribution of prediction residuals is shown in a Q-Q plot for each layer (shifted horizontally for clarity) with the black lines showing the theoretical relationship, and the overall distribution histogram is shown on the right. %Especially in early layers, the prediction residuals are approximately $N(0,1)$ distributed, validating that the uncertainties are reasonably well calibrated across the range of resolutions.
}
 	\label{fig:calibration}
\end{figure}

As we describe in the following section, the RBF kernel in each layer is trained to maximize the marginal likelihood of the data that it sees, and thereby minimize the discrepancy with the underlying generating distribution $f^{(\ell+1)}$. 
While this measurement/projection approach is effective in many scenarios, in networks with many layers or a very large number of observations uncertainty information can get attenuated as it passes through the layers, a phenomenon which we investigate in appendix \ref{sec:pathologies}.
With a network that is trained on a version of MNIST that
is randomly subsampled to 75 pixels, in \figref{fig:calibration} (left) 
we evaluate the mean and uncertainty of the internal feature maps as we vary the number the number of pixels of the inputs at test time. As expected, the mean functions for the feature maps slowly converge and the predicted uncertainties decrease in magnitude as the input resolution is increased. 
In \figref{fig:calibration} (middle) we show that in early layers the uncertainties decrease at a similar rate to the $O(1/\sqrt{N})$ of discretization error that we would expect from a standard convolutional layer which is discretized to a square grid.\footnote{A 2D discrete convolution layer using $N=m^2$ points can be interpreted as a Riemann sum approximation of the continuous integral and will therefore have an error of $O(1/m)=O(1/\sqrt{N})$ which is the same rate as would be achieved through Monte Carlo sampling.}
Despite the fact that these resolutions differ substantially from those seen at training time and the fact that there are no explicit uncertainty targets for these internal layers, the predictions are reasonably well calibrated as demonstrated in \figref{fig:calibration} (right). While the prediction residuals have fatter tails than a standard Gaussian, the mean and standard deviation are close to the theoretically optimal $0$ and $1$ values across a range of resolutions. %We note that these uncertainties are less calibrated in the later layers, a phenomenon which we investigate in appendix \ref{sec:pathologies}.

\subsection{Training procedure} %\marc{Possible candidate for moving to the appendix}\MW{Maybe compress it but we should leave the objective in the paper I think}

Our neural network has two sets of parameters: the channel mixing and diffusion parameters, 
$\{(M^{(\ell)},\mW^{(\ell)},\bm \beta^{(\ell)},\bm \Sigma^{(\ell)})\}_{\ell=1}^L$, as well as kernel hyperparameters of the Gaussian Processes $\{(l^{(\ell)},a^{(\ell)})\}_{\ell=1}^L$.
%\subsection{Training procedure}
% We have described in the previous how to efficiently approximate the neural network map $\Phi(f)$ on a GP $f$ interpolating the input data. (Recall the definition $\Phi = {\cal P}
%     \circ 
%     \sigma 
%     \circ 
%     {\cal A}^{(L)}
%     \cdots 
%     \circ 
%     \sigma
%     \circ 
%     {\cal A}^{(2)}
%     \circ 
%     \sigma
%     \circ 
%     {\cal A}^{(1)}$.)
% We now use the mean of that process, $
%     \mathbb{E}_{f\sim {\cal GP}}
%     [\Phi(f)]
% $,
% as the logits used for a classification problem. The covariance
% $\mathbb{E}_{f\sim {\cal GP}}
%     [\Phi(f) \Phi(f)^T]$
% encodes the propagation of the input uncertainty through the network. It results in a prediction uncertainty due to the scarcity of samples of the input signal.
% The more information about the input we have, the smaller the input uncertainty will be and consequently the smaller the output uncertainty should be, as in the limit of infinite resolution the input is a deterministic quantity.
% To compute the statistics of the feature vector $\Phi(f)$ we face however the following challenges: 1) applying the linear operator ${\cal A}$ of eq.~\eqref{eq:A_general} cannot be done in general in closed form; 2) even if we can apply in closed form ${\cal A}$, computing the statistics of $\Phi(f)$, where $f$ is a GP, is challenging due to the non-linearities in $\Phi$. 
% Below we shall present solutions to both problems.
We train all parameters jointly on the loss $L_{\mathrm{task}} + \lambda L_{\mathcal{GP}}$, where $L_{\mathrm{task}}$ is the cross entropy with logits given by the mean $\mu_P$ of the pooled features ${\cal P}(f^{(L)}) \sim {\cal N}(\mu_P, \Sigma_P)$ and $L_{\mathcal{GP}}$ is the marginal log likelihoods of the GP feature maps:
\begin{equation}
    L_{\mathcal{GP}}(f) = \frac{1}{2}
    \sum_{\ell=1}^L \sum_{\alpha=1}^{c_\ell}\bigg[{\bigg(\bm{f}^T_\alpha [K_{XX}+S_\alpha]^{-1}\bm{f}_\alpha\bigg)} + \log \det {[K_{XX}+S_\alpha]}+N\log{2\pi}\bigg]^{(\ell)}
    \,,
\end{equation}
where for each layer $\ell$, $\bm{f}_\alpha = \big[f_\alpha(x_1),...,f_\alpha(x_N)\big] \in \mathbb{R}^{N}$ are the observed values for channel $\alpha$ at locations $X = [x_1,\dots,x_N]$, $K_{XX}$ is the covariance of the RBF kernel and $S_\alpha=\mathrm{diag}(\sigma^2_\alpha)$ the measurement noise for each channel $\alpha$ and spatial location. Notably the GP marginal likelihood is independent of the class labels.%does not depend on the labels which means that is has a regularizing effect that can be computed even on test data.
% \subsection{Adaptive Meshing}
% We can now choose the observation locations $\{x_i^{(\ell)}\}_{i=1}^m$ at each layer in a way so as to minimize the variance of the predictions or simply to maximize the Marginal likelihood of the GP. This can be done on both the training data and the testing data without making use of the labels. For every example input to the network, we add additional trainable parameters for the observation locations for each layer in the network which we initialize to sampling locations of the original continuous signal and optimize during training.% $\sum_i\Tr(\mathcal{A}k_p\mathcal{A}^\dagger)(x_i,x_i)$. It's possible that this computation may be done in closed form when $k$ is an RBF kernel, but it's not obvious.
% %Since one may need to use numerical methods to optimize this (locally computed) quantity, one could instead minimize the sampling locations over the uncertainty in the output logits of the network.

\section{Experimental Results}
We evaluate the Probabilistic Numeric CNN on two different problems which have incomplete and irregular observations. 

\textbf{Superpixel MNIST} is an adaptation of the MNIST dataset where the $784$ pixels of the original images are replaced by $75$ salient \emph{superpixels} that are non uniformly spread throughout the domain and are different for each image \citep{monti2017geometric}. Despite the simplicity of the underlying images, the lack of grid structure and high fraction of missing values make this a challenging task. Example inputs are visualized at the left of \figref{fig:network}. We compare to Monet \citep{monti2017geometric}, SplineCNN \citep{fey2018splinecnn}, Graph Convolutional Gaussian Processes (GCGP) \citep{walker2019graph}, and Graph Attention Networks (GAT) \citep{avelar2020superpixel}.
% For more details on the dataset and experimental setup, see \ref{sec:datasets}. 
As shown in table \ref{table:superpixel}, the probabilistic numeric CNN greatly outperforms the competing methods reducing the classification error rate by more than $3\times$ over the previous state of the art.

\begin{table}[h]
    \centering
    \label{table:superpixel}
    \mbox{%\small 
    \begin{tabular}{l c}
    %\toprule
     MNIST Superpixel 75 & Error Rate ($\downarrow$)\\
     \midrule
      Monet & 8.89\\
      SplineCNN & 4.78\\
      GCGP  & 4.2\\
      GAT  & 3.81\\
      PNCNN & \textbf{1.24}$\pm$0.12\\
      PNCNN w/o $\sigma$ & 3.03$\pm$0.10\\
    \end{tabular}
    \centering
    \label{table:physionet}
    \begin{tabular}{l cc}
    %\toprule
     PhysIONet2012 & AP ($\uparrow$) & AUROC ($\uparrow$)\\
     \midrule
      IP-Nets& 51.86 & 86.24\\
      SEFT-ATTN & 53.67 & 85.14 \\
      GRU-D & \textbf{54.97} & \textbf{86.99} \\
      PNCNN & 53.92$\pm$.17 & 86.13$\pm$.06\\
    \end{tabular}
    }
    \caption{Left: Classification Error on the 75-SuperPixel MNIST problem. 
    Right: Average Precision and Area Under ROC curve metrics for PhysioNET2012. Mean and standard deviation are computed over $3$ trials.}
\end{table}

We conduct an ablation study where uncertainty propagation is removed: the probabilistic ReLU is replaced with the deterministic one applied to the mean, and the uncertainties in each layer are set to $0$. This form of the network still makes use of the fact that the input function is continuous by use of the GP interpolation of the means, but crucially it does not integrate the uncertainty in the computation resulting from the missing data. While this variant (PNCNN w/o $\sigma$) with a $3.03\%$ error rate outperforms existing methods from the literature, it is substantially worse than the PNCNN that integrates the uncertainty at $1.24\%$ error. This validates both that the underlying architecture (using the continuous convolution operators) has good inductive biases and that reasoning about discretization errors probabilistically can improve performance directly.

\begin{figure}
\centering
\includegraphics[width=.9\linewidth]{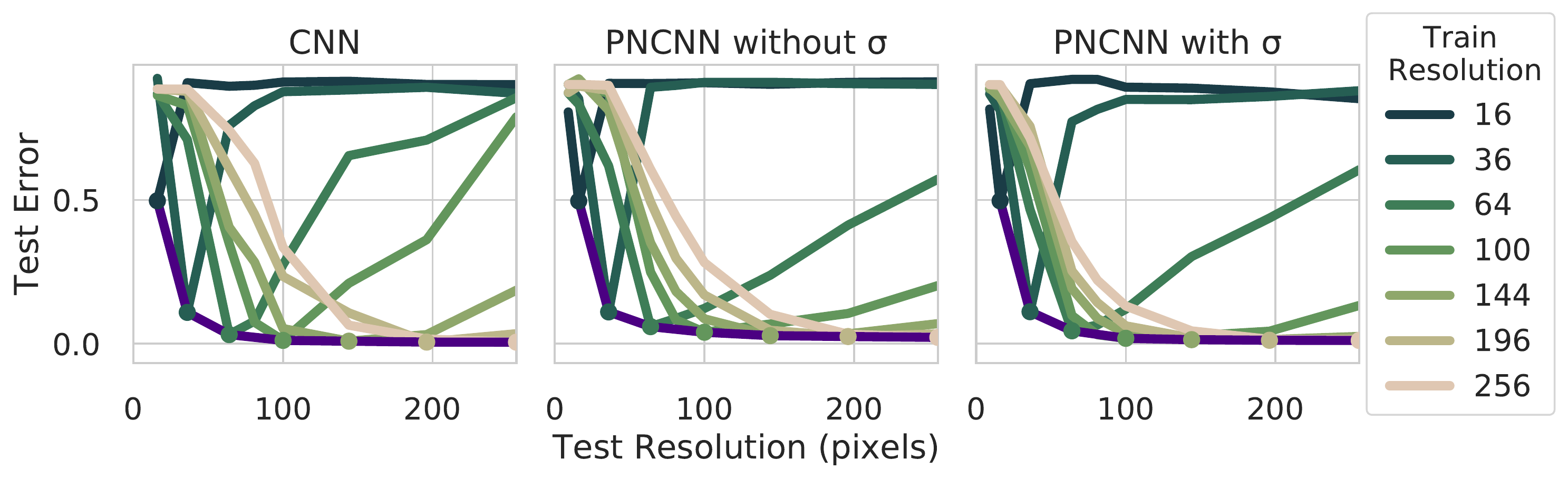}
\caption{Zero shot generalization to other resolutions: Having trained on MNIST at a given training resolution shown by the color, we evaluate the performance of an PNCNN, PNCNN without uncertainty, and an ordinary CNN on varying test resolutions. Notably, PNCNN with uncertainty is the most robust.}
 	\label{fig:train_test_shift}
\end{figure}

In \figref{fig:train_test_shift} we evaluate the performance on zero shot generalization to a different test resolution for a variety of training resolutions. In order to compare to an ordinary CNN we sample MNIST on a regular square grid: PNCNN with uncertainty is the most robust to this train test sampling distribution shift, followed by PNCNN w/o uncertainty, and finally the ordinary CNN which is quite sensitive to these changes.

\textbf{Irregularly Spaced Time Series}
For the second task, we evaluate our model on the irregularly spaced time series dataset PhysioNet2012 \citep{silva2012predicting} for predicting mortality from ICU vitals signs. This dataset is particularly challenging because different vital sign channels are observed at different times, even within a single patient record. This means that we cannot compute the GP inference formula of \eqref{eq:posterior} efficiently for all channels simultaneously because the observation points $\{x_i\}$ and hence the matrices $K$ in that formula differ between the channels, increasing computational complexity. To circumvent this difficulty, we employ a stochastic diagonal estimator to compute the variances as described in appendix \ref{sec:diag_estimator}. We compare against IP-Nets \citep{shukla2019interpolation},
      SEFT-ATTN \citep{horn2019set}, and GRU-D \citep{che2018recurrent} as reported in \citet{horn2019set}. PNCNN performs competitively, although not a breakout performance as in the image dataset which we attribute to the use of the stochastic variance estimates over an exact calculation.

\section{Conclusion}

Based on the ideas of probabilistic numerics, we have introduced a new class of neural networks which model missing values and discretization errors probabilistically \emph{within the layers} of a CNN. The layers of our network are a series of operators defined on \emph{continuous} functions, removing dependence on shortcut features like the sampling locations and distribution. On irregularly sampled and incomplete spatial data we show improved generalization and robustness. 

As a closing comment, we note that, owing to exchangeability of finite distributions of a stochastic process, our architecture is permutation equivariant. We therefore envision new applications of our framework to graph data in the future.
Data on curved manifolds will be described by gauge equivariant GPs and PDEs, generalizing considerably the mathematical models of this work.

\bibliography{iclr2020_conference}
\bibliographystyle{plainnat}
\newpage
\appendix

\section{Review of Gaussian Processes}
\label{sec:GP_review}
We briefly review here the main ideas of Gaussian Processes for machine learning, see \cite{stein2012interpolation,rasmussen2006gaussian} for more details.
We start to explain how to use stochastic processes for Bayesian inference. We see the stochastic process as prior over functions $p(f)$ and as we are given samples $\bm{x} = (x_1,\dots,x_N), \bm{y} = (y_1,\dots,y_N), y_i\equiv f(x_i)$, we update our beliefs about the function by constructing the posterior via Bayes rule: $p(f | \bm{y}, \bm{x}) = p(\bm{y}|f,\bm{x}) p(f) / p(\bm{y}|\bm{x})$. Here we need to specify the likelihood of the data with our model $p(\bm{y}|f,\bm{x})=\prod_{i=1}^N p(y_i|f,x_i)$ and the denominator, called the evidence or marginal likelihood, follows: $p(\bm{y}|\bm{x})=\mathbb{E}_{f\sim p(f)}[p(\bm{y}|f,\bm{x})]$.
The power of this approach is that the value of the signal $y$ at an unseen point $x$ has an uncertainty which depends on our knowledge of its neighbourhood: $p(y|x,\bm{y},\bm{x}) = \mathbb{E}_{f\sim p(f|\bm{y},\bm{x})}[p(y|f,x)]$ and allows us to reason probabilistically about the underlying signal.

A particular convenient class of random function is Gaussian processes (GPs) for which inference can be done exactly. A stochastic process can be presented in terms of the finite distributions of the random variables $\{f(x_i)\}_{i=1}^M$ at points $\{ x_i \}_{i=1}^M$. For a GP these distributions are Gaussian and can be defined uniquely by specifying means and covariances, and so a GP is specified entirely by its mean function $\mu(x)$ and covariance kernel $k(x,x')$. We shall write $f\sim {\cal GP}(\mu, k)$.
Let us assume a Gaussian likelihood model as well, i.e.~
$p(y_i|f,x_i) = {\cal N}(f(x_i), \sigma_i^2)$, where $\sigma_n$ represents aleatoric uncertainty on the measurement. (For simplicity we take here the function to be scalar valued but the reasoning can be easily generalized.)
Then properties of the Gaussian distribution (see \cite[Chap. 2]{rasmussen2006gaussian} for a detailed derivation of the formulas) lead to the following posterior distributions after seeing data $\bm{y}, \bm{x}$: $p(f|\bm{y},\bm{x}) = {\cal GP}(\mu_p, k_p)$, with
\begin{align}
\mu_p(x)=\bm{k}(x)^T [K + S]^{-1} \bm{y}
\,,\quad
k_p(x,x')
=
k(x, x')
-
\bm{k}(x)^T [K + S]^{-1}
\bm{k}(x')\,.
\end{align}
where $K_{ij}=k(x_i,x_j), k(x)_i = k(x, x_i)$ and $S = \mathrm{diag}(\sigma_i^2)$.
%The predictive distribution is then $p(y|x,\mY,\mX) = {\cal N}(\mu_p(x), k_p(x,x) + \sigma_n^2)$.

\section{From discrete to continuous convolutional layers}
\label{sec:disc_cont_conv}
We here show that the general formula
\begin{align}
\label{eq:A_general} 
{\cal A}
=
\sum_k W_k {\rm e}^{{\cal D}_k}
\end{align}
with ${\cal D}_k$ a function of spatial derivatives, 
reduces in the case of discrete input space ${\cal X}$ to the usual convolution we encounter in deep learning.

For simplicity we shall assume a $1$d grid as input space ${\cal X} = \{1,\dots,N\}$.
Let us start by recalling the form of the classical discrete convolution when $C_{\ell}=C_{\ell+1}=1$. We define a convolutional layer as a linear map that commutes with the translation operator. To make the symmetry exact, we need assume periodic boundaries. Then in the standard basis of $\mathbb{R}^N$, $\{e_i\}_{i=1}^N$ of vectors localized at site $i$, the translation operator $\tau$ acts as $\tau e_i = e_{i+1 \mod N}$. An $N\times N$ matrix $B$ is translation invariant iff $\tau B = B \tau$. Since $\tau$ is diagonal in Fourier space, the most general solution is $B=F \text{diag}(\hat{\bm{b}}) F^{-1}$, where $F_{jk} = {\rm e}^{\frac{2\pi i}{N} jk}$ is the discrete Fourier transform. Such matrices are called circulant and can be written alternatively as $B = \sum_{i=0}^{N-1} b_{N-i} \tau^i$, $\bm{b} = F \hat{\bm{b}}$. Explicitly:
\begin{equation}
B=
\begin{pmatrix}
b_0     & b_{1} & \dots  & b_{N-2} & b_{N-1}  \\
b_{N-1} & b_0    & b_{1} &         & b_{N-2}  \\
\vdots  & b_{N-1}& b_0    & \ddots  & \vdots   \\
b_{2}  &        & \ddots & \ddots  & b_{1}   \\
b_{1}  & b_{2} & \dots  & b_{N-1} & b_0 \\
\end{pmatrix}\,.
\end{equation}
This shows that the most general convolutional layer is a circulant matrix. E.g.~if $b_i=0$ unless $i=0,1,N-1$, $B$ coincides with the matrix representing a periodic convolution of filter size $3$.
The matrix $B$ is invertible as long as $\hat{b}_k\neq 0$ for all $k$. In a convolutional network the parameters $b_i$ are random variables and the measure of the set where $B$ is not invertible is zero.
Thus the role of ${\rm e}^{{\cal D}}$ is replaced in the discrete case by the $B$.

%We are now ready to discuss how the discrete analog of \eqref{eq:A_general}. 
The discrete analog of ${\cal A}$ is then:
\begin{align}
    \label{eq:A_discrete}
    A
    =
    \sum_i 
    W_i
    \otimes 
    B_i
    \,.
\end{align}
Introducing the unit matrices $E_{\alpha,\beta}$ which have $1$ at the row $\alpha$ and column $\beta$ and $0$ otherwise, we can rewrite it as:
\begin{align}
    \label{eq:A_discrete}
    A
    =
    \sum_{j,\alpha,\beta} 
    E_{\alpha,\beta}\otimes \tau^j
    W^{\alpha,\beta}_j
    \,,\quad
    W^{\alpha,\beta}_j
    =
    \sum_i
    W^{\alpha,\beta}_i
    b_{i,N-j}
    \,.
\end{align}
Since $E_{\alpha,\beta}\otimes \tau^j$ is a linear basis of the space of convolutional layers, we see that \eqref{eq:A_general} indeed reduces to the usual one when discretizing the input domain and is a principled generalization to the continuous domain.

% \section{Comparison of differential and exponential operators}
% \label{sec:linear_op_comparison}

% \begin{figure}
% \begin{multicols}{2}
% \centering
% \includegraphics[width=\linewidth]{figs/derivative_vs_pde_features.png}\\

% \includegraphics[width=\linewidth]{figs/derivative_vs_pde_features.png}\\

% \end{multicols}
% \caption{\textbf{Left:} Time evolution of diffusion equation for 1d and 2d signals.\marc{TODO} \textbf{Right:} Comparison of differential operator and exponentiated operator features.}
%  	\label{fig:linear_op}
% \end{figure}

% In figure \ref{fig:linear_op} we show the action of a diffusion layer on 2d signals.
% (The exact procedure to compute such outputs is explained below in section \ref{sec:rbf_conv}.)
% This has to be compared with linear operators that are linear combinations of derivatives, such as $\mathcal{D} = \beta^\top \nabla + (1/2)\nabla^\top\Sigma\nabla$, which are inherently \textit{local}, depending only on the values of the function in an infinitesimal neighborhood of $x$, and therefore have receptive field of $0$. \footnote{In the presence of a fine enough sampling, features built up through derivative operators and pointwise functions like ReLUs will have no interactions between input values that are separated by a finite distance, which we make precise and demonstrate in Appendix \ref{}\marc{TODO}.}

\section{Greens Function}\label{subsec:greens}

Given the operator $\mathcal{D} = \beta^\top \nabla +\tfrac{1}{2}\nabla^\top\Sigma\nabla$ we can compute the action of ${\rm e}^{t\mathcal{D}}$ in terms of convolutions. Using the $d$ dimensional Fourier transforms $\mathcal{F}[h](k) = (2\pi)^{-\frac{d}{2}}\int h(x)e^{-ik^\top x}dx$ and $\mathcal{F}^{-1}=\mathcal{F}^\dagger$, we can rewrite the derivative operator $\mathcal{D}$ in terms of elementwise multiplication in the Fourier domain, which diagonalizes $\mathcal{D}$. Since $\nabla= \mathcal{F}^{-1} (ik)
\mathcal{F}$,
\begin{equation}
    \mathcal{D} = \mathcal{F}^{-1}(i\beta^\top k-\tfrac{1}{2}k^\top\Sigma k)\mathcal{F}.
\end{equation}
Using the series definition  ${\rm e}^{t\mathcal{D}} = \sum_{n=0}^\infty (t\mathcal{D})^n/n!$, we have:
\begin{equation}
    e^{t\mathcal{D}} = \mathcal{F}^{-1}e^{t(i\beta^\top k-\tfrac{1}{2}k^\top\Sigma k)}\mathcal{F}.
\end{equation}
Applying this operator to a test function $h(x)$ yields
\begin{equation}\label{eq:greens_application}
    {\rm e}^{t\mathcal{D}}h = \mathcal{F}^{-1}[e^{t(i\beta^\top k-\tfrac{1}{2}k^\top\Sigma k)}\mathcal{F}[h](k)] = \mathcal{F}^{-1}[\mathcal{F}[G_t]\cdot\mathcal{F}[h]] = G_t*h,
\end{equation}
where the final step follows from the Fourier convolution theorem, and we define the function $G_t = \mathcal{F}^{-1}[{\rm e}^{t(i\beta^\top k-\tfrac{1}{2}k^\top\Sigma k)}]$. Directly applying the Fourier integral yields a Gaussian integral
\begin{equation}
    G_t(x)=(2\pi)^{-\frac{d}{2}}\int {\rm e}^{ik^\top(x+t\beta) - \tfrac{1}{2}k^\top t\Sigma k}dk = {\rm e}^{-\frac{1}{2}(x+t\beta)^\top(t\Sigma)^{-1}(x+t\beta)}\mathrm{det}(2\pi t\Sigma)^{-1/2}.
\end{equation}
This function $G_t(x) = \mathcal{N}(x;-t\beta,t\Sigma)$ is nothing but a multivariate heat kernel, the Greens function (also known as the fundamental solution or time propagator) for the diffusion equation
$\partial_tG_t(x-x') = \mathcal{D}G_t(x-x')$, and indeed $\lim_{t\to 0}G_t(x-x') = \delta(x-x')$.
\section{Integral Pooling}\label{sec:integral_pooling}

%\MW{Move expressions to appendix, just briefly mention what you in words}
The integral pooling operator $\mathcal{P}[f] = \int_{\mathbb{R}^d} f(x) dx$ can be applied to the Gaussian process just like any other linear operator. Given $f^{(L)} \sim \mathcal{GP}(\mu,k)$, we have that
\begin{equation}
    \mathcal{P}f^{(L)} \sim  \mathcal{GP}(\mathcal{P}\mu,\mathcal{P}k\mathcal{P}')=\mathcal{N}(\mathcal{P}\mu,\mathcal{P}k\mathcal{P}').
\end{equation}
Again, computing the mean $\mu_P = \mathcal{P}\mu$ and covariance matrix $\Sigma_P = \mathcal{P}k\mathcal{P}'$ we need just to be able to apply $\mathcal{P}$ to the RBF kernel.
\begin{align}
    \mathcal{P}k_{\mathrm{RBF}}(x') &= \int_{\mathbb{R}^d}k_{\mathrm{RBF}}(x,x')dx = a\\
    \mathcal{P}k_{\mathrm{RBF}}\mathcal{P}' &= \int_{\mathbb{R}^d\times \mathbb{R}^d}k_{\mathrm{RBF}}(x,x')dxdx' = \infty
\end{align}
For many applications such as image classification using the mean logit value, we require only the predictive mean, so an unbounded covariance matrix $\Sigma_P$ is acceptable. We use this form for all of our experiments.

However for some applications an output uncertainty can be useful, so we also provide a variant that integrates over a finite region $[0,1]^d$, $\mathcal{P}f = \int_{[0,1]^d} f(x)dx$.
\begin{align}
    \mathcal{P}k_{\mathrm{RBF}}(x') &= \int_{[0,1]^d}k_{\mathrm{RBF}}(x,x')dx = a\prod_{i=1}^d\big[\Phi(\tfrac{x_i'}{\ell})-\Phi(\tfrac{x_i'-1}{\ell})\big]\\
    \mathcal{P}k_{\mathrm{RBF}}\mathcal{P}' &= \int_{[0,1]^d\times [0,1]^d}k_{\mathrm{RBF}}(x,x')dxdx' =a\big[ \ell\sqrt{\tfrac{2}{\pi}}(e^{-1/2\ell^2}-1) + 2\Phi(\tfrac{1}{\ell})-1\big]^d
\end{align}
where $\Phi$ is again the univariate standard normal CDF.

\section{Equivariance}
\label{sec:equivariance_app}
\subsection{Related Work}
We note that there has been considerable research effort in the development of equivariant CNNs which we build on top of. The group equivariant CNN was introduced by \citet{cohen2016group} for discrete groups on lattices. This work has been extended for continuous groups \citep{worrall2017harmonic,zhou2017oriented} and with steerable equivariance \citep{cohen2016steerable,weiler2019general} where other group representations are used. There have also been group equivariant networks designed for point clouds and other irregularly spaced data \citep{thomas2018tensor,finzi2020generalizing,fuchs2020se,de2020gauge}. In \citet{shen2020pdo}, layers using finite difference estimation of derivative operators are used for defining equivariant layers in an equivariant CNN, effectively a change of basis.

Most closely related to our PDE operator approach to equivariance is work by \citet{smets2020pde}. In this work, the authors define layers of their convolutional network through the time evolution of a PDE which is a nonlinear generalization of the diffusion equation, which includes pooling like behaviour. The PDEs explored \citet{smets2020pde} are equivariant by choice of the parameters in the PDE, and when incorporating multiple channels are very similar to enforcing equivariance on the operator $\mathcal{A} = \sum_kW_ke^{\mathcal{D}_k}$ which we investigate below and include it as a special case. 

However, as we note, enforcing equivariances in this form of $\mathcal{A}$ outside of translation leads to only solutions without many degrees of freedom which is also mentioned in \citet{smets2020pde}. We prove that this approach to incorporating multiple channels and equivariance is very limited and leads only to trivial isotropic solutions (even with nontrivial feature representations) in Section \ref{sec:trivial_solns}. We instead provide an alternate approach based on intrinsically coupled systems of PDEs over the channels $\mathcal{A} = e^{\sum_kW_k\mathcal{D}_k}$ which does not have this deficiency. We also derive the general conditions for equivariance of linear operators.

% In this section we extend the theory of steerable equivariance from convolutions to linear operators more generally.
% \marc{TODO: Introduce related work on equivariance.}
% Derivatives and Equivariance:
% PDO-eConvs: Partial Differential Operator Based Equivariant Convolutions,
% \citep{shen2020pdo}
% Equivariance:
% G-CNN,
% Steerable CNNs,
% LieConv,
% GEM-CNN,
% GAT-Superpixel \citep{avelar2020superpixel}
% PDE-based Group Equivariant Convolutional Neural Networks

\subsection{Translation Equivariance}
A key factor in the generalization of convolutional neural networks is their translation equivariance. Patterns in different parts of an input signal can be seen in the same way because convolution is translation equivariant. Our learnable linear operators $\mathcal{A}$ are equivariant to continuous transformations. Two linear operators $e^\mathcal{C}$ and ${\rm e}^\mathcal{B}$ commute $[{\rm e}^\mathcal{C},{\rm e}^\mathcal{B}]=0$ if and only if their generators commute: $[\mathcal{C},\mathcal{B}]=0$. Since the generator of diffusions $\mathcal{D}_i$ is a sum of derivative operators, and the generators of translations are just $\nabla$ as mentioned in section \ref{subsec:continuous_convolution}, the two commute: $[\mathcal{D}_k,\nabla]=0$. Therefore $[\mathcal{A},\tau_a] = [\sum_kW_k{\rm e}^{\mathcal{D}_k},\tau_a] = \sum_kW_k[{\rm e}^{\mathcal{D}_k},{\rm e}^{a^\top\nabla}]=0$ and $\mathcal{A}$ is translation equivariant.

\subsection{Steerable Equivariance for Linear Operators}
For some tasks like medical segmentation, aerial imaging, and chemical property prediction there are additional symmetries in the data it makes sense to exploit other than mere translation equivariance. Below we show how to enforce equivariance of the Linear operator $\mathcal{A}$ to other symmetry groups $G$ such as the group of continuous rotations $\mathrm{SO}(d)$ in $\mathbb{R}^d$. Applying equivariance constraints separately on each of the components of ${\rm e}^{\mathcal{D}_i}$ on top of translation equivariance yields very restricted set of operators. For example, enforcing equivariance to continuous rotations $G = \mathrm{SO}(d)$, the operator must be an isotropic heat kernel: $\mathcal{D}_k=c_k\nabla^\top\nabla$. The reason for this apparent restriction is a result of considering the different channels independently, as scalar fields. 

The alternative is to use features fields which transform under more general representations of the symmetry group, introduced in steerable-CNNs \citep{cohen2016steerable} and used in \citep{worrall2017harmonic,thomas2018tensor,weiler2018learning,weiler2019general} and others. In this way, the symmetry transformation acts not only on the spatial domain $\mathcal{X}$, but also transforms the channels. The way that the group acts on $\mathbb{R}^c$ (i.e.~the channels) is formalized by a representation matrix $\rho(g) \in \mathbb{R}^{c\times c}$ for each element $g\in G$ in the transformation group that satisfies $\forall g,h \in G: \rho(gh) = \rho(g)\rho(h)$. Choosing the type of each intermediate feature map is equivalent to choosing their representations, and we describe a simple way of doing this with tensor representations in the later section.

% In the following section, we will use the summation notation where repeated indices are assumed to be summed over to reduce clutter. Given the operator $\mathcal{A}$ that acts on a function $f$ by $\mathcal{A}^{\alpha\beta}f^\beta = W^{\alpha\beta}_\gamma \mathcal{D}_\gamma f^\beta$, we will replace each of output channels, input channels, and mixing channels (denoted with the indices $\alpha,\beta,\gamma$) with the components for a certain composite representation $U_{out}$,$U_{in}$,$U_{mix}$, according to some chosen collection of tensor ranks.

\textbf{Operator Equivariance Constraint}:
Returning to linear operators, we derive the equivariance constraint and show how to use constructs from the previous sections to implement steerable rotation equivariance.  Equivariance of a linear operator $\mathcal{A}: (\mathbb{R}^d\to\mathbb{R}^{c_{in}}) \to (\mathbb{R}^d\to\mathbb{R}^{c_{out}})$ requires that, for any input function, transforming the input function first (both argument and channels) and applying $\mathcal{A}$ is equivalent to first applying $\mathcal{A}$ and then transforming the output: $\mathcal{A}\rho_{in}(g)L_{g}f = \rho_{out}(g)L_{g}\mathcal{A}f$ where $L_gf(x) = f(g^{-1}x)$. Rearranging the terms, one sees that the equivariance constraint on the linear operator $\mathcal{A}$ is:
\begin{equation}\label{eq:op_constraint}
    \rho_{out}(g)L_g\mathcal{A}L_{g^{-1}}\rho_{in}(g^{-1}) = \mathcal{A},
\end{equation}
where the operators $L_g$ and $L_g^{-1}$ are understood not to act on the representation matrices $\rho$ (although implicitly a function of $g$). As shown in Appendix \ref{sec:convolution_constraint}, eq.~\ref{eq:op_constraint} is a direct generalization of the equivariance constraint for convolutions $\forall x: \rho_{out}(g)K(g^{-1}x)\rho_{in}(g^{-1}) =K(x)$ described in the literature \citep{weiler2019general,cohen2019general}.

%This means that \ref{eq:op_constraint} can be rewritten as
As shown in Appendix \ref{sec:trivial_solns}, the equivariance constraint for continuous rotations applied to the diffusion operators $\mathcal{A} = \sum_k W_k {\rm e}^{\mathcal{D}_k}$ has only the trivial solutions of isotropic diffusion without any drift. For this reason we instead consider a more general form of diffusion operator where the PDE itself couples the different channels. For the coupled PDE:
\begin{equation}
    \frac{\partial f}{\partial t} = \sum_k W_k \mathcal{D}_k f
\end{equation}
the time evolution contains the matrices $W_k$ \emph{in} the exponential
$\mathcal{A} = e^{\sum_k W_k \mathcal{D}_k}$. Like with the example of translation above, this operator is equivariant if and only if the infinitesmal generator $\sum_kW_k\mathcal{D}_k$ is equivariant.

Because equation \ref{eq:op_constraint} applies generally to linear operators and not just convolutions, we can compute the equivariance constraint for these derivative operators. We can simplify the summation $\sum_k W_k \mathcal{D}_k = \sum_k W_k (\beta_k^T\nabla+(1/2)\nabla^T\Sigma_k\nabla)$ by writing it in terms of the collection of matrices $B_i = \sum_k W_k \beta_{ki}$ and $S_{ij}=(1/2)\sum_k W_k \Sigma_{kij}$ to express $\mathcal{A}_{\mathrm{deriv}} = \sum_i B_i\partial_i + \sum_{i,j}S_{ij}\partial_i\partial_j$ where the indices $i,j=1,2...,d$ enumerate the spatial dimensions of each vector $\beta_k$ and each matrix $\Sigma_k$. As we derive in appendix \ref{sec:matrix_exp}, the necessary and sufficient conditions for the equivariance of $\sum_k W_k \mathcal{D}_k$ and therefore $\mathcal{A}$ is that
$\forall g\in G: [\rho_{out} \otimes \rho_{in}^* \otimes \rho_{(1,0)}](g)\mathrm{vec}(B) = \mathrm{vec}(B)$ and $\forall g\in G: [\rho_{out} \otimes \rho_{in}^*\otimes \rho_{(2,0)}](g)\mathrm{vec}(S)=\mathrm{vec}(S)$ where $\mathrm{vec}(\cdot)$ denotes flattening the elements into a single vector and $\rho_{(r,s)}$ is the tensor representation with $r$ covariant and $s$ contravariant indices. 
%As detailed in appendix \ref{sec:nullspace}, we can then solve numerically for equivariant subspace of $B$ and $S$ that satisfy the constraints with a QR decomposition.

\subsection{Generalization of Equivariance Constraint for Convolutions}\label{sec:convolution_constraint}
This equivariance constraint is a direct generalization of the equivariance constraint for convolution kernels as described in \citet{weiler2019general,cohen2019general}. In fact, when $\mathcal{A}$ is a \textit{convolution} operator, $\mathcal{A}f = K*f$, the action of $L_g$ by conjugation $\mathcal{A}$ is equivalent to transforming the argument of the kernel $K$:
\begin{align*}
    L_g(K*)L_{g^{-1}}f(x) &= \int K(g^{-1}x-x')f(gx')d\mu(x') \\
    &= \int K(g^{-1}(x-x''))f(x'')d\mu(x'') = (L_g[K])*f.
\end{align*}

Letting both sides of eq \ref{eq:op_constraint} act on the product of a constant unit vector $e_i$ and a delta function, $f=e_i\delta$ the expression $\forall e_i: \rho_{out}(g)L_g[K]\rho_{in}(g^{-1})*e_i\delta=K *e_i\delta$ can be rewritten as $\forall x: \rho_{out}(g)K(g^{-1}x)\rho_{in}(g^{-1}) =K(x)$ which is precisely the constraint for steerable equivariance for convolution described in the literature. \footnote{This assumes as is typically done that measure $\mu$ over which the convolution is performed is left invariant. For the more general case, see the discussion in \cite{bekkers2019b}.}

\subsection{Equivariant Diffusions with Matrix Exponential}\label{sec:matrix_exp}
Below we solve for the \textbf{necessary} and \textbf{sufficient} conditions for the equivariance of the operator $\mathcal{A}_{\mathrm{deriv}}$.

We will use tensor representations for their convenience, but the approach is general to allow other kinds of representations. A rank $(p,q)$ tensor $t$ is an element of the vector space $T_{(p,q)}:=V^{\otimes p}\otimes (V^*)^{\otimes q}$ where $V$ is some underlying vector space, $V^*$ is its dual and $(\cdot)^{\otimes p}$ is the tensor product iterated $p$ times. In common language $T_{(0,0)}$ are scalars, $T_{(1,0)}$ are vectors, and $T_{(1,1)}$ are matrices. Given the action of a group $G$ on the vector space $V$, the representation on $T_{(p,q)}$ is $\rho_{(p,q)}(g) = g^{\otimes p} \otimes (g^{-\top})^{\otimes q}$ where $-\top$ is inverse transpose and $\otimes$ on the matrices is the tensor product (Kronecker product) of matrices. Composite representations can be formed by stacking different tensor ranks together, such as a representation of $50$ scalars, $25$ vectors, $10$ matrices and $5$ higher order tensors: $T_{(0,0)}^{50}\oplus T_{(1,0)}^{25}\oplus T_{(1,1)}^{10}\oplus T_{(1,2)}^{5}$, where $\oplus$ in this context is the same as the Cartesian product. For a composite representation $U=\bigoplus_i T_{(p_i,q_i)}$ the group representation is similarly $\rho_U(g) =\bigoplus_i \rho_{(p_i,q_i)}(g)$ where $\oplus$ concatenates matrices as blocks on the diagonal.

Noting that the operator $L_g$ that acts only on the argument and the matrix $\rho_{in}(g)$ acts only on the components, the two commute and we can rewrite the constraint for $\mathcal{A}_\mathrm{deriv}$ as 
\begin{equation}\label{eq:deriv_constraint}
    \sum_i\rho_{out}(g)B_i\rho_{in}(g^{-1})L_g\partial_iL_{g^{-1}}+\sum_{ij}\rho_{out}(g)S_{ij}\rho_{in}(g^{-1})L_g\partial_i\partial_jL_{g^{-1}} = \mathcal{A}_{\mathrm{deriv}}
\end{equation}
We can simplify the expression $L_g\partial_iL_{g^{-1}}$ by seeing how it acts on a function. For any differentiable function $\partial_iL_{g^{-1}}f(x) = \frac{\partial}{\partial x_i}[f(gx)] = \sum_{j}g_{ji}[\partial_jf](gx) = L_{g^{-1}} \sum_{j}g_{ji}\partial_j f(x)$ where $g_{ij}$ are the components of the matrix $g$. Since this holds for any $f$, we find that $L_g\nabla L_{g^{-1}} = g^T\nabla$ and therefore $L_g\nabla\nabla^T L_{g^{-1}} = L_g\nabla L_{g^{-1}}L_g\nabla^T L_{g^{-1}} = g^T\nabla \nabla^Tg$.

Since equation \ref{eq:deriv_constraint} holds as an operator equation, it must be true separately for each component $\partial_i$ and $\partial_i\partial_j$. This means that the constraint separates into a constraint for $B$ and a constraint for $S$:
\begin{enumerate}
    \item $\forall g,i: \sum_jg_{ij}\rho_{out}(g)B_j\rho_{in}(g^{-1}) = B_i$
    \item $\forall g,i,j: \sum_{kl}g_{i\ell}g_{ik}\rho_{out}(g)S_{\ell k}\rho_{in}(g^{-1}) = S_{ij}$.
\end{enumerate}
%However, we can apply the more general form of the constraint in eq \ref{eq:op_constraint} to more general linear operators like derivatives $\mathcal{A} = \sum_\gamma W_\gamma \mathcal{D}_\gamma$. 
% Noting that the operator $L_g$ that acts only on the argument and the matrix $\rho_{in}(g)$ acts only on the components, the two commute and we can rewrite the constraint for these derivative operators as
% \begin{equation}
%     \sum_\gamma \rho_{out}(g)W_\gamma \rho_{in}(g^{-1}) L_{g}\mathcal{D}_\gamma L_{g^{-1}} = \sum_\gamma W_\gamma \mathcal{D}_\gamma.
% \end{equation}
% Inserting the identity $I = \rho_{mix}(g^{-1})\rho_{mix}(g)$:
% \begin{equation}
%     \sum_{c\gamma d}\bigg(\rho_{out}(g)W_c \rho_{in}(g^{-1})\rho_{mix}(g^{-1})_{c \gamma}\bigg)\bigg(\rho_{mix}(g)_{\gamma d} L_{g}\mathcal{D}_\gamma L_{g^{-1}}\bigg) = \sum_\gamma W_\gamma \mathcal{D}_\gamma.
% \end{equation}

These relationships can be expressed more succinctly by flattening the elements of $B$ and $S$ into vectors: $[\rho_{out}(g) \otimes \rho_{in}(g^{-T}) \otimes \rho_{(1,0)}(g)]\mathrm{vec}(B) = \mathrm{vec}(B)$ and $[\rho_{out}(g) \otimes \rho_{in}(g^{-T}) \otimes \rho_{(2,0)}(g)]\mathrm{vec}(S)=\mathrm{vec}(S)$. 
%As detailed in appendix \ref{sec:nullspace}, we can then solve numerically for equivariant subspace of $B$ and $S$ that satisfy the constraints with a QR decomposition.

\subsection{Rotation Equivariance Constraint for Scalar Diffusions has only Trivial Solutions}\label{sec:trivial_solns}
%\RB{not checked}
The diffusion operator $\mathcal{A} = \sum_k W_ke^{\mathcal{D}_k}$ leads to only trivial $\beta_k=0$ and $\Sigma_k\propto I$ if it satisfies the continuous rotation equivariance constraint.

\textbf{Proof:}

%We can alternatively enforce the equivariance constraint on the greens function for $e^{\mathcal{D}_k}$ using existing methodology from the standard steerable CNNs literature.
The application of $e^{\mathcal{D}_k}$ is just a convolution with the Greens function
\begin{equation}
    \sum_k W_ke^{\mathcal{D}_k}f = \sum_k W_k[e^{-\frac{1}{2}(x+\beta_k)^\top\Sigma_k^{-1}(x+\beta_k)}\mathrm{det}(2\pi \Sigma_k)^{-1/2}]*f = \sum_kW_kG_k*f
\end{equation}
where the Greens function is the multivariate Gaussian density: $G_k(x)=\mathcal{N}(x;-\beta_k,\Sigma_k)$.

As shown in appendix \ref{sec:convolution_constraint}, for convolutions the operator constraint is equivalent to the kernel equivariance constraint $\rho_{out}(g)K(g^{-1}x)\rho_{in}(g^{-1}) =K(x)$ from \citep{weiler2019general}. With $K(x) = \sum_k W_k G_k(x)$ this reads:
\begin{equation*}
    \forall x\in \mathbb{R}^d,g\in G:\quad \sum_{k}\rho_{out}(g)W_k\mathcal{N}(g^{-1}x;-\beta_k,\Sigma_k)\rho_{in}(g^{-1}) =\sum_{k} W_k\mathcal{N}(x;-\beta_k,\Sigma_k),
\end{equation*}
For rotations $g\in \mathrm{SO}(2)$ where we can parametrize $g_\theta = e^{\theta J}$ in terms of the antisymmetric matrix $J=[[0,1],[-1,0]] \in \mathbb{R}^{2\times 2}$ and the translation operator can be written $L_g = e^{-\theta x^TJ^T\nabla}$, we can take derivatives with respect to $\theta$ to get (now with double sums implicit):
\begin{equation*}
    \forall x\in\mathbb{R}^d: \sum_k\big[d\rho_{out}W_k\mathcal{N}(x;-\beta_k,\Sigma_k) - W_k\mathcal{N}(x;-\beta_k,\Sigma_k)d\rho_{in} - W_k(x^T J^T\nabla)\mathcal{N}(x;-\beta_k,\Sigma_k)\big] = 0.
\end{equation*}
Here the Lie Algebra representation of $J$ is $d\rho:=\frac{\partial}{\partial \theta} \rho(g_\theta)|_{\theta=0}$. Factoring out the normal density:
\begin{equation*}
    \forall x\in\mathbb{R}^d:\quad\sum_k\big[d\rho_{out}W_k - W_kd\rho_{in} - W_k(x^T J^T\Sigma^{-1}_k(x+\beta_k))\big]\mathcal{N}(x;-\beta_k,\Sigma_k) = 0.
\end{equation*}

Without loss of generality we may assume that each of the Gaussians $\beta_k,\Sigma_k$ pairs are distinct since if they were not then we could replace the collection with a single element. Since the (finite) sum of distinct Gaussian densities is never a Gaussian density, and monomials of order $>0$ multiplied by a Gaussian density cannot be formed with sums of Gaussian densities or sums multiplied by monomials of a different order and Gaussian densities are never $0$, this constraint separates out into several independent constraints.
\begin{enumerate}
    \item $\forall i: d\rho_{out}W_k = W_kd\rho_{in}$
    \item $\forall i,x: W_k (x^TJ^T\Sigma_k^{-1}\beta_k) = 0$
    \item $\forall i,x: W_k (x^TJ^T\Sigma_k^{-1}x) =0$
\end{enumerate}

We may assume w.l.o.g. that $W_k$ is not $0$ for all components of the matrix (otherwise we could have deleted this element of $k$ and continue). Therefore there is some component which is nonzero, and the expressions in parentheses in equations $2$ and $3$ must be $0$. Given that this holds for all $x$, eq $3$ implies:
$J^T\Sigma_k^{-1} = 0$ or equivalently $\Sigma_k^{-1}J=0$ because $\Sigma_k$ is symmetric, and since $J=-J^T$ this can be expressed concisely as $[\Sigma_k^{-1},J]=0$ for which the only symmetric solution is proportional to the identity $\Sigma_k = c_k I$. Since both $\Sigma_k$ and $J$ are invertible, equation $2$ yields $\beta_k=0$. Therefore there are no nontrivial solutions for $\beta,\Sigma$ in $\mathcal{A} = \sum_k W_ke^{\mathcal{D}_k}$ for continuous rotation equivariance.

\section{Dataset and Training Details}
In this section we elaborate on some of the details regarding hyperparameters, network architecture, and the datasets.

As described in the main text, the PNCNN is composed of a chain of convolutional blocks containing a convolution layer, a probabilistic ReLUs, and linear channel mixing layer (analogue of the colloquial $1\times 1$ convolution). In each of these convolutional blocks, the input is a collection of points and feature mean value at those points along with the feature elementwise standard deviation at those points: $\{(x_i,\mu(x_i),\sigma(x_i)\}_{i=1}^N$. These observations seed the GP layer, and the block is evaluated at the same collection of points for the output (although it can be evaluated elsewhere since it is a continuous process, and we make use this fact to visualize the features in figures \ref{fig:network} and \ref{fig:calibration}).

\textbf{Hyperparameters:}
For the PNCNN on the Superpixel MNIST dataset, we use $4$ PNCNN convolution blocks with $c=128$ channels and with $K=9$ basis elements for the different drift and diffusion parameters in $\sum_{k=1}^KW_ke^{\mathcal{D}_k}$. We train for $20$ epochs using the Adam optimizer \citep{kingma2014adam} with $\mathrm{lr}=3 10^{-3}$ with batch size $50$.

For the PNCNN on the PhysioNet2012 dataset, we use the variant of the PNCNN convolution layer that uses the stochastic diagonal estimator described in appendix \ref{sec:diag_estimator} with $P=20$ probes. In the convolution blocks we use $c=96$ channels, $K=5$ basis elements and we train for $10$ epochs using the same optimizer settings above. For both datasets we tuned hyperparameters on a validation set of size $10\%$ before folding the validation set back into the training set for the final runs. Both models take about 2 hours to train.

\textbf{SuperPixel-MNIST}
We source the SuperPixel MNIST dataset \citep{monti2017geometric} from \citet{FeyLenssen2019} consisting of $60k$ training examples and $10k$ test represented as collections of positions and grayscale values $\{(x_i,f(x_i))\}_{i=1}^{75}$ at the $N=75$ super pixel centroids.

\textbf{PhysioNet2012}
We follow the data preprocessing from \citet{horn2019set} and the $10k$-$2k$ train test split. The individual data points consist of $42$ irregularly spaced vital sign time series signals as well as $5$ static variables: Gender, ICU Type, Age, Height, Weight. We use one hot embeddings for the first two categoric variables, and we treat each of these static signals as fully observed constant time series signals.  As the binary classification task exhibits a strong label imbalance, $14\%$ positive signals, we apply an inverse frequency weighting of $1/.14$ to the binary cross entropy loss.

\section{Stochastic Diagonal Estimation for PhysioNet2012}\label{sec:diag_estimator}

In order to compute the mean and variance of the rectified Gaussian process, the activations of the probabilistic ReLU, we need compute the diagonal of $\mathcal{A}k_p\mathcal{A}'(x_n,x_n)$ for the relevant points $\{x_n\}_{n=1}^N$.

In the usual case where each of the channels $\alpha=1,2,...,c$ are observed at the same locations this can be done efficiently. First one computes the application of $e^{\mathcal{D}_i}$ on the left and $e^{\mathcal{D}_j'}$ on the right onto the posterior $k_p$:
\begin{equation*}
    N_{ij} = (e^{\mathcal{D}_i}k_pe^{\mathcal{D}_j'})(x_n,x_n) = (e^{\mathcal{D}_i}ke^{\mathcal{D}_j'})(x_n,x_n)-(e^{\mathcal{D}_i}\mathbf{k}^\top)(x_n)[K+S]^{-1}(\mathbf{k}e^{\mathcal{D}_j'})(x_n)
\end{equation*} where $k$ is the RBF kernel and we have reused the notation from appendix \ref{sec:GP_review}. Notably, this quantity is the same for each of the channels, and the elementwise variance is just:
\begin{equation}
    v_\alpha(x_n) = (\mathcal{A}k_p\mathcal{A}')_{\alpha\alpha}(x_n,x_n) = \sum_{i,j,\beta} W^{\alpha\beta}_iN_{ij}W^{\alpha\beta}_j
\end{equation}
where the $\alpha,\beta$ index the channels of each of the matrices $W_i$. Because $N$ is the same for all channels, we can compute this quantity efficiently with a reasonable memory cost and compute.

For the PhysioNet2012 dataset where the observation points differ between the channels we must consider a different observation set $\{x_n^\beta\}_{n=1}^N$ for each channel $\beta$. This means that evaluated kernel depends on the channel and we have the objects: $\mathbf{k}^{\beta}$, $K^\beta$ and $S^\beta$. As a result, we have an additional index for $N_{ij}^\beta$ and the desired computation is
\begin{equation}\label{eq:direct}
    v_\alpha(x_n^\alpha) = (\mathcal{A}k_p\mathcal{A}')_{\alpha\alpha}(x_n^\alpha,x_n^\alpha) = \sum_{i,j,\beta} W^{\alpha\beta}_iN_{ij}^\beta W^{\alpha\beta}_j.
\end{equation}
While each of the terms in the computation can be computed without much difficulty, performing the summation explicitly requires an unreasonably large memory cost and also compute.

However, by the same approach we can consider the full covariance matrix $\mathbf{B}_{(\alpha n)(\beta m)} = (\mathcal{A}k_p\mathcal{A}')_{\alpha\beta}(x_n^\alpha,x_m^\beta)$, and while it would not be feasible to compute this matrix directly we \textit{can} define matrix vector multiplies onto vectors of size $\mathbb{R}^{cN}$ implicitly using the sequence of operations that define it. Crucially, this sequence of operations has much more modest memory consumption (and compute cost) over the direct expression in \eqref{eq:direct}. These implicit matrix vector multiplies can then be used to compute a stochastic diagonal estimator \citep{bekas2007estimator} given by:
\begin{equation}
    \hat{v}_\alpha(x_n^\alpha) = \tfrac{1}{P}\sum_{p=1}^P z_p\odot \mathbf{B}z_p
\end{equation}
with Gaussian probe vectors $z_p \sim \mathcal{N}(0,I)$, and where $\odot$ is elementwise multiplication (see \citet{bekas2007estimator} for more details on this stochastic diagonal estimator). We use this estimator with $P=20$ probes for computing the variances for PhysioNet. We note that with $P=20$ the variance estimates are still quite noisy, however without the estimator cannot readily apply the PNCNN to PhysioNet. We leave a better approach for handling this kind of data to future work.

\section{Pathologies in Projection to RBF Gaussian Process}\label{sec:pathologies}
In section \ref{sec:projection} describe an approach by which a Gaussian process with a complex mean and covariance function is \textit{projected} down to the posterior of a (simpler) RBF kernel GP from a set of observations. We know given the representation capacity of the RBF kernel that with the right set of observations, a complex function can be well approximated in principle. However, the relationship for uncertainty is less straightforward.

The properties of the input Gaussian process must be conveyed to the output Gaussian process by only the (uncorrelated) noisy observations $\{(x_i,\mu(x_i),\sigma(x_i))\}_{i=1}^N$. As the uncertainty in original GP increases, so do the measurement uncertainties in the transmission, and therefore the output GP also has a higher uncertainty. However, the uncertainty in the input GP is in the form of a full covariance kernel $k(x,x')$ and it seems that individual observations will not easily be able to communicate the $\textit{covariance}$ of the values of the GP function at different spatial locations despite the heterogeneous noise model.

Fundamentally, the problem is that the observation values are treated as independent, an incorrect assumption which has other knock-on effects when the number of observations is large. With some fixed measurement error no matter how high but a large enough set of independent observations, the mean value can be pinned down precisely. If in contrast the observations are not independent, then there may be a situation where the mean value cannot be known more precisely than some limiting uncertainty. This effect leads the output GP to have less uncertainty and be more confident in the values that it should be given the input GP.

If the observations are sparse, then the effective sample size of the estimator for the mean of the GP at any given location is small, and then the amount by which uncertainty is underestimated is small. However, if there are many many observations then this kind of observation transmission of information with the independence assumption will attenuate the uncertainty. We would also expect that over the course of many layers, this attenuation can accumulate. We believe that this is what causes the poorer uncertainty calibration in layers 3 and 4 of the PNCNN shown in figure \ref{fig:calibration}. We hope that this problem can be resolved perhaps by removing the independence assumption or providing an alternative projection method in future work.

\section{Central Limit Theorem for Stochastic Processes}\label{sec:clt}

We derive a variant a variant of the Lyapunov central limit theorem (CLT) holding for stochastic processes. The main ideas is that the result for processes follows from applying the multivariate Lyapunov CLT to the joint distribution of each finite collection of values as per the definition of a Gaussian process.

In more details the argument goes as follows.
We are given $C$ stochastic processes $g_c(x)$ and we assume that they are weakly dependent, i.e.~$\mathbb{E}[ g_c(x) g_{c'}(x') ]\to \mathbb{E}[ g_c(x)] \mathbb{E}[g_{c'}(x') ]$ as $|c-c'|\gg 1$, for any $x,x'$. 
We would like to show that $\bar{g}(x)\sim {\cal GP}(\mu, k)$, where $\mu(x) = \sum_{c=1}^C g_c(x)$ and $k(x,x') =\sum_{c,c'=1}^C \mathbb{E}[g_c(x)g_{c'}(x')]$.
For these formulas to make sense, we need some bounds on the moments of $g_c$. If the individual components $g_c$ scale as $1/\sqrt{C}$, then the covariance if finite.

Now choose any finite collection of indices $x_{1},x_{2},\dots,x_N$. Then consider the random vector $\bar{g}(x_i) = \sum_{c=1}^C g_c(x_i)$, $i=1,\dots,N$. We can now apply the CLT to deduce that $\{ \bar{g}(x_i) \}_{i=1}^N$ is Gaussian distributed. Since a stochastic process is determined by its finite distributions, we can conclude that the random function $\bar{g}(x) \to {\cal GP}(\mu, k)$, as was to be shown.

%\newpage
\section{Moments of rectified Gaussian random variables} 
\label{sec:relu_covar}

Let $\bm{f} \sim {\cal N}(\bm{\mu}, \bm{\Sigma})$ be a $d$ dimensional random Gaussian vector. We compute here
\begin{align}
    \mathbb{E}(\text{ReLU}(f_1) \cdots \text{ReLU}(f_d))
    &=
    \frac{1}{N_\Sigma}
    \int_{\bm{f} > 0} {\rm d}^d \bm{f} 
    \, f_1 \cdots f_d
    \exp -\tfrac{1}{2} (\bm{f}-\bm{\mu})^T \mSigma^{-1} (\bm{f}-\bm{\mu})\\
    N_\mSigma &=
    (2\pi)^{d/2} \det(\mSigma)^{1/2}
    \,.
\end{align}
We use the generating function technique. Define
\begin{align}
    Z(\bm{b})
    &=
    \frac{1}{N_\Sigma}
    \int_{\bm{f} > 0} {\rm d}^d \bm{f} 
    \exp [-\tfrac{1}{2} (\bm{f}-\bm{\mu})^T \mSigma^{-1} (\bm{f}-\bm{\mu}) + \bm{b}^T \bm{f}] \\
    &=
    \frac{1}{N_\Sigma}
    e^{\bm{b}^T \bm{\mu}}
    \int_{\bm{f} > -\bm{\mu}} {\rm d}^d \bm{f} 
    \exp [-\tfrac{1}{2} \bm{f}^T \mSigma^{-1} \bm{f} + \bm{b}^T \bm{f}]
    \,.
\end{align}
Then
\begin{align}
    \mathbb{E}(\text{ReLU}(f_1) \cdots \text{ReLU}(f_d))
    &=
    \frac{\partial}{\partial b_1}\cdots \frac{\partial}{\partial b_d}
    Z(\bm{b}) \bigg\rvert_{\bm{b} = 0}
    \,.
\end{align}
To compute $Z(\bm{b})$ we proceed as in Gaussian case. We change variables to
\begin{align}
    \bm{f} = \mSigma \bm{b} + \bm{g}\,,
\end{align}
and define $\bm{z} = \bm{\mu} + \mSigma \bm{b}$ to get:
\begin{align}
    Z(\bm{b})
    &=
    e^{\bm{b}^T \bm{\mu} + \tfrac{1}{2}\bm{b}^T\mSigma \bm{b}}
    \frac{1}{N_\mSigma}
    \int_{\bm{g} < +\bm{z}} 
    {\rm d}^d \bm{g} 
    \exp [-\tfrac{1}{2} \bm{g}^T \mSigma^{-1} \bm{g}]\\
    &=
    e^{S(\bm{b})} 
    \,\Phi^{(d)}(\bm{z}; 0, \mSigma)
    \,,
    \quad
    S(\bm{b})
    =
    \bm{b}^T \bm{\mu} + \tfrac{1}{2}\bm{b}^T\mSigma \bm{b}\,.
\end{align}
$\Phi$ being the multivariate standard Normal CDF:
\begin{align}
    \Phi^{(d)}(\bm{z}; \bm{\mu}, \mSigma)
    &=
    \int_{\bm{g} < +\bm{z}} 
    {\rm d}^d \bm{g} 
    \psi^{(d)}(\bm{g}; \bm{\mu}, \mSigma)
    \,,\\
    \psi^{(d)}(\bm{g}; \bm{\mu}, \mSigma)
    &=
    \frac{1}{N_\mSigma}
    \exp [-\tfrac{1}{2} (\bm{g}-\bm{\mu})^T \mSigma^{-1} (\bm{g}-\bm{\mu})]
    \,.
\end{align}
Now we compute the first two derivatives. Note that in $d=1$, denoting $\sigma^2 = \Sigma$:
\begin{align}
    \frac{\partial}{\partial z} \Phi^{(1)}(z, 0, \sigma^2)
    =
    \psi^{(1)}(z, 0, \sigma^2)
    \,.
\end{align}
In $d=2$, we can use the conditional probability decomposition to get the required derivatives:
\begin{align}
    \psi^{(2)}(\bm{g}; 0, \mSigma)
    &=
    \psi^{(1)}(g_1; \alpha_1 g_2,\beta_1)
    \cdot
    \psi^{(1)}(g_2; 0, \Sigma_{22})\\
    \alpha_1 &= \Sigma_{12}\Sigma_{22}^{-1}\,,\beta_1 = \Sigma_{11} - \Sigma_{12} \Sigma_{22}^{-1} \Sigma_{21}\\
    \partial_{z_2} \Phi^{(2)}(\bm{z}, 0, \mSigma)
    &=
    \psi^{(1)}(z_2; 0, \Sigma_{22})    
    \int_{-\infty}^{z_1} 
    {\rm d}g_1
    \psi^{(1)}(g_1; \alpha_1 z_2,\beta_1)\,, \\
    &=
    \psi^{(1)}(z_2; 0, \Sigma_{22})    
    \Phi^{(1)}(z_1, \alpha_1 z_2, \beta_1)
    \\
    \partial^2_{z_2} \Phi^{(2)}(\bm{z}, 0, \mSigma)
    &=
    -\frac{z_2}{\Sigma_{22}}
    \psi^{(1)}(z_2; 0, \Sigma_{22})    
    \Phi^{(1)}(z_1, \alpha_1 z_2, \beta_1)\\
    &\quad+
    \psi^{(1)}(z_2; 0, \Sigma_{22})    
    \partial_{z_2}\Phi^{(1)}(z_1, \alpha_1 z_2, \beta_1)
    \\
    \partial_{z_1} \partial_{z_2} \Phi^{(2)}(\bm{z}, 0, \mSigma)
    &=
    \psi^{(2)}(\bm{z}; 0, \mSigma) 
    \,.
\end{align}

So denoting $\partial_i = \frac{\partial}{\partial b_i}$, we get:
\begin{align}
    \partial_iZ(\bm{b})
    &=
    (\mu_i + \sum_j \Sigma_{ij} b_j) Z(\bm{b})
    +
    \underbrace{
        e^{S(\bm{b})}
        \sum_\ell \partial_{z_\ell}
        \Phi^{(d)}(\bm{z}, 0, \mSigma)
        \Sigma_{\ell,i}
        }_{m_i(\bm{b})}
    \\
    \partial_k\partial_iZ(\bm{b})
    &=
    \Sigma_{ik} Z(\bm{b})
    +
    (\mu_i + \sum_j \Sigma_{ij} b_j) \partial_k Z(\bm{b})  \\
    &+
    (\mu_k + \sum_j \Sigma_{kj} b_j) m_i(\bm{b})
    +
    e^{S(\bm{b})}
    \sum_{\ell,q} \partial_{z_\ell} \partial_{z_q} 
    \Phi^{(d)}(\bm{z}, 0, \mSigma)
    \Sigma_{\ell,i} \Sigma_{q,k}
    \,.
\end{align}
In particular, for the first moment $d=1$ we have:
\begin{align}
    \mathbb{E}(\text{ReLU}(f)) = \mu \Phi^{(1)}(\mu; 0, \sigma^2) + \psi^{(1)}(\mu, 0, \sigma^2) \sigma^2
    \,,
\end{align}
which coincides with \eqref{eq:rectified_moments_mu}. Note that $\psi^{(1)}(\mu, 0, \sigma^2) = \tfrac{1}{\sigma}\psi(\mu/\sigma;0,1)$ because of the normalization factor. For the second moments $d=2$ we have:
\begin{align}
    \mathbb{E}(\text{ReLU}(f_1)\text{ReLU}(f_2)) 
    &= 
    \Sigma_{12} \Phi^{(2)}(\bm{\mu}; 0, \mSigma)
    +
    \mu_1 \mu_2 \Phi^{(2)}(\bm{\mu}; 0, \mSigma)
    + 
    \mu_1 m_2(0)
    + 
    \mu_2 m_1(0)\\
    &\qquad 
    + 
    \sum_{\ell,q=1,2} 
    \Sigma_{\ell,1} \Sigma_{q,2}
    \partial_{z_\ell} \partial_{z_q} 
    \Phi^{(2)}(\bm{z}, 0, \mSigma)|_{\bm{b}=0}
    \,.
\end{align}
which can be rewritten in the form of \eqref{eq:rectified_moments_Sigma}.
% Note that the second moment at coinciding points simplifies to
% \begin{align}
%     \mathbb{E}[\sigma(f(x))^2]&=(\mu^2+\sigma^) \Phi^{(1)}(\mu; 0, \Sigma)+\mu\psi^{(1)}(\mu, 0, \Sigma)\Sigma
%     \,.
% \end{align}
% \RB{simplify the covariance?}

\end{document}